\documentclass[3p, authoryear]{elsarticle}
\usepackage[utf8]{inputenc}
\usepackage{amssymb,amsfonts,amsmath}
\usepackage{graphicx}
\usepackage{multirow}
\usepackage{lipsum}
\usepackage[kernelfbox]{xcolor}
\usepackage{pgfplots,pgfplotstable}
\usepackage{pgfkeys}
\usepackage{float}
\usepackage{wrapfig}
\restylefloat{figure}
\usepackage{booktabs}
\usepackage[english]{babel}

\usepackage{caption}
\usepackage[flushleft]{threeparttable}
\usepackage{subcaption}
\usepackage{footnote}
\makesavenoteenv{tabular}
\makesavenoteenv{table}
\usepackage{scrextend}

\usepackage{setspace}
\usepackage{natbib}
\usepackage{array}
\usepackage{calc}
\DeclareMathOperator*{\argmin}{arg\,min~}
\DeclareMathOperator*{\argmax}{arg\,max~}

\pgfplotsset{
	invoke before crossref tikzpicture={\tikzexternaldisable},
	invoke after crossref tikzpicture={\tikzexternalenable},
}

\usepgfplotslibrary{groupplots} 
\usetikzlibrary{matrix}
\usetikzlibrary{patterns}
\usepgfplotslibrary{external} 
\usepgfplotslibrary{units}
\usepgfplotslibrary{dateplot} 
\pgfplotsset{compat=newest}
\usetikzlibrary{patterns}

\definecolor{Turquoise}{RGB}{141, 211, 199}
\definecolor{darkgreen}{RGB}{0,100,0}

\newcommand*{\regularbox}[1]{\color{black}% open a group for a local setting
   \setlength{\fboxsep}{-1\fboxrule}% the rule will be inside the box boundary
   \fbox{\hspace{1.1pt}\textbf{\strut#1}\hspace{1.2pt}}% print the box, with some padding at the left and right
\color{black}  }% close the group

\usepackage[colorlinks]{hyperref}
\usepackage{chngcntr}

\usepackage[nameinlink,capitalize]{cleveref}
\crefformat{appendix}{#2#1#3}

\newcommand{\x}{\mathbf{x}}
\newcommand{\y}{\mathbf{y}}
\newcommand{\z}{\mathbf{z}}

\newcommand{\s}{\mathbf{s}}

\newcommand{\m}{\mathbf{m}}

\newcommand{\h}{\mathbf{h}}
\newcommand{\w}{\mathbf{w}}
\newcommand{\E}{\mathbf{E}}

\newcommand{\U}{\mathbf{U}}
\newcommand{\V}{\mathbf{V}}
\newcommand{\W}{\mathbf{W}}

\journal{Computer Speech \& Language}
\begin{document}

\begin{frontmatter}
\title{Online Learning for Effort Reduction in \\
Interactive	Neural Machine Translation}

%% Group authors per affiliation:
\author[]{\'Alvaro Peris\corref{corauth}}
\ead{lvapeab@prhlt.upv.es}

\author[]{Francisco Casacuberta}
\ead{fcn@prhlt.upv.es}

\address{Pattern Recognition and Human Language Technology Research Center,\\ Universitat Polit\`ecnica de Val\`encia, \\
        Camino de Vera s/n, 46022 Valencia, SPAIN }
\cortext[corauth]{Corresponding author:  
Phone: (34) 96 387 70 69}

\begin{abstract}

Neural machine translation systems require large amounts of training data and resources. Even with this, the quality of the translations may be insufficient for some users or domains. In such cases, the output of the system must be revised by a human agent. This can be done in a post-editing stage or following an interactive machine translation protocol.

We explore the incremental update of neural machine translation systems during the post-editing or interactive translation processes. Such modifications aim to incorporate the new knowledge, from the edited sentences, into the translation system. Updates to the model are performed on-the-fly, as sentences are corrected, via online learning techniques. In addition, we implement a novel interactive, adaptive system, able to react to single-character interactions. This system greatly reduces the human effort required for obtaining high-quality translations.

In order to stress our proposals, we conduct exhaustive experiments varying the amount and type of data available for training. Results show that online learning effectively achieves the objective of reducing the human effort required during the post-editing or the interactive machine translation stages. Moreover, these adaptive systems also perform well in scenarios with scarce resources. We show that a neural machine translation system can be rapidly adapted to a specific domain, exclusively by means of online learning techniques.
\end{abstract}


\end{frontmatter}

\section{Introduction}

In the last years, the field of machine translation (MT) has witnessed impressive progresses, mostly due to the advances achieved in corpus-based machine translation. Nowadays, MT systems are useful tools for many users and companies, automatically providing translations of acceptable quality in many cases~\citep{Crego16,Wu16}.

Nevertheless, we are still far from solving the MT problem~\citep{Koehn17}. The systems still produce wrong translations, that may be intolerable for some users or domains. For example, translations of medical records must be accurate and error-free. Moreover, the translation problem has many subtleties that make it hard for machines to tackle it: discourse adequacy, anaphora resolution, domain-specific meanings, stylistic forms, etc. 

In scenarios that require high-quality translations, the outputs of the systems are usually reviewed by a human agent, who corrects the errors made by the MT system. Thus, the user is benefited by the use of MT, since it allows to achieve higher productivity than translating from scratch \citep{Arenas08,Green13b}. The process of correcting MT hypotheses is known as post-editing. 

Given the advantages of post-editing, new approaches to efficiently produce good translations emerged, aiming to increase the productivity of this process, thereby diminishing the human effort required. Among them, interactive machine translation (IMT) \citep{Barrachina09,Casacuberta09,Foster97} is one of the most attractive strategies.

The IMT framework introduces the human corrector into the editing process: the system reacts to each human action. As the user makes a correction, the system has more information for generating an alternative translation, hopefully better than the previous one, which spares effort to the human translator.

With translation post-editing or IMT, we obtain high-quality translations. These translations contain new knowledge, which is prone to be profited by an adaptive MT system, taking advantage of these new samples and adapting its models to them. Online learning (OL) methods are suitable for this goal.
OL is a machine learning paradigm in which data is available sequentially and models are updated incrementally, sample to sample. The online learning framework can be structured according to four main stages~\citep{Murphy12}:

\begin{enumerate}
	\label{enum:ol}	
	\item \label{item:ol_x} A sample is presented to the system.
	\item \label{item:ol_prediction} The system provides a prediction for this sample.
	\item \label{item:ol_correct-label} The correct prediction is provided to the system.
	\item \label{item:ol_learning} The system uses the correct prediction to adapt its models.
\end{enumerate}

Therefore, in the OL framework, there is no distinction between the training and prediction phases: the system is continuously learning and predicting, as data become available. 
Therefore, the MT systems can be retrained as the correction process goes on, avoiding to make the same errors again and being adapted to a given domain or tailored to the style of the human corrector.

The MT technology has recently experienced a revolution. During several years, phrase-based statistical machine translation (PB-SMT) models~\citep{Koehn03} were the state-of-the-art in MT. But in the last years, a novel corpus-based technique emerged: the so-called neural machine translation (NMT), in which the translations are generated solely by neural networks. NMT achieves more fluent and natural translations~\citep{Wu16} than previous PB-SMT systems. Moreover, NMT systems perform exceptionally well under an IMT paradigm, as shown by \citet{Knowles16} and \citet{Peris17a}.

But the NMT technology also has weaknesses. As studied by \citet{Koehn17}, NMT systems have lower quality when translating out-of-domain sentences and they require larger amounts of training data than PB systems. Given these findings, NMT faces a dilemma: on the one hand, training with large amounts of in-domain data may be infeasible, due economic restrictions or to the lack of domain-specific data. On the other hand, we need to train with large amounts of data, in order to obtain good translations. A possible solution for this issue relies on the domain adaptation field \citep{Ben10}. In this scenario, a model trained on a large corpus of out-of-domain samples, aims to perform well on a different domain. In the field of MT, under the domain adaptation umbrella, we find a large set of techniques \citep{Axelrod11,Chen17,Farajian17}.

In this work, we aim to leverage the aforementioned issues of NMT systems. Our goal is to adapt translation systems on-the-fly, during the post-editing or IMT stages. For doing this, we take advantage of the human-revised translations generated in the post-editing or IMT processes and apply OL techniques to the NMT system. Additionally, we deepen into the application of OL into IMT framework, using the NMT technology. To the best of our knowledge, this is the first work that puts together interactive neural machine translation and online learning. We thoroughly evaluate our models, setting up three different scenarios, that account for the casuistic that can happen in an industrial MT setting. We show that OL can be effectively used for enhancing the NMT system and reducing the human effort required during the correction process. Our main contributions are:
\begin{enumerate} 
\item We study the application of OL techniques in the post-editing and IMT scenarios, using NMT systems. 
\item We introduce a simple and effective way for performing character-level interactions in a (sub)word-based interactive NMT (INMT) system.
\item We conduct a large experimentation, using public corpora containing different features from varied domains. We stress the translation systems, applying them in three different translation scenarios.
\item We show that OL brings significant improvements to the translation engines, in terms of translation quality and human effort reduction. Comparisons with other works in the literature also show that our adaptive, interactive systems are able to outperform the existing state-of-the-art.
\item We open-source all code developed in this work, in order to make research reproducible.
\end{enumerate}
%

% Paper organization
The rest of this manuscript is structured as follows: the related work is reviewed in~\cref{sec:related-work}. Next, \cref{sec:NMT} briefly introduces the NMT technology, while the interactive protocol for NMT is presented in \cref{sec:IMT}. OL is described and applied together with INMT in \cref{sec:OL}. \cref{sec:ExperimentalFramework} describes the experimental setup of this work. Results are presented and discussed in \cref{sec:Results}. Finally, we conclude the work and trace future lines of work in \cref{sec:Conclusions}.

\section{Related work}
\label{sec:related-work}
This work puts together three thoroughly studied fields: neural machine translation, interactive machine translation and online learning. In this section, we briefly review the progress made in the last years in each one of these fields.

%Neural MT
\subsection{Neural machine translation}

Although the first attempts of performing machine translation with neural networks date from long ago \citep{Castano97,Forcada97}, NMT only took off recently. 
\citet{Kalchbrenner13} reintroduced full neural MT, although the results were non-competitive with respect to classical PB-SMT systems. Nevertheless, in the next year,~\citet{Cho14} and \citet{Sutskever14} proposed two similar sequence-to-sequence models, applied to the MT problem with encouraging results. These works were based on an encoder--decoder architecture, implemented with recurrent neural networks, with long short-term memory (LSTM) units~\citep{Hochreiter97} or gated recurrent units (GRU)~\citep{Cho14}. From here, the NMT technology had a meteoric trajectory. \citet{Bahdanau15} introduced the so-called attention model in the sequence-to-sequence framework. This allowed the system to selectively focus on parts of the input sequence, providing good results when modeling long sequences. %Alternative attention models have bee proposed~\citep{Luong15a}, but the original one seems to be the most effective, according to recent investigations~\citep{Britz17}.

This attentional NMT system was the basis of many works, which aimed to tackle its main issues, namely the management of large vocabularies and the out-of-vocabulary problem~\citep{Jean15,Luong15b}. The most satisfactory solution, was the use of subword sequences instead of words~\citep{Luong16,Sennrich16}. This has become a de facto standard in NMT. 

NMT systems are typically trained by means of stochastic gradient descent (SGD), with a maximum likelihood objective (see \cref{sec:NMT}). Nevertheless, some works explored alternative cost functions. Reinforcement learning \citep{Shen16} or minimum risk training \citep{Wu16} strategies have been applied to NMT systems, generally with positive results. 

At this moment, the NMT technology has reached the translation industry, and some companies have already adopted it as translation engine~\citep{Crego16,Wu16}.

Moreover, the encoder--decoder framework can be applied to many other problems apart from MT, generally with good results. Among these applications we can find image captioning~\citep{Xu15}, video captioning~\citep{Yao15,Peris16}, parsing~\citep{Vinyals15b} or speech translation~\citep{Duong16}. Multi-task learning approaches take advantage from this versatility and are also obtaining encouraging results~\citep{Kaiser17}.

\subsection{Interactive machine translation}

Since \citet{Foster97} introduced the IMT, this approach has been continuously revised and developed \citep{Alabau13,Barrachina09,Casacuberta09,Langlais02,Macklovitch05}, demonstrating and improving its capabilities. Hence, the original IMT protocol has been extended and modified in several ways: \citep{Alabau11,Sanchis08} included multimodal feedback, \citep{Azadi15,Cai13,Green14} improved the suffix generation, \citep{Gonzalez10b} integrated of confidence measures in the interactive pipeline, etc. 

Given the recent success of NMT, this technology has also be adapted to fit into the interactive framework~\citep{Knowles16,Peris17a}. Alternative technologies, such as translation memories, also were modified to allow interaction~\citep{Green14}. Other works aimed to build resource-agnostic IMT systems \citep{Perez14}.

A significant effort has also been spent in overcoming the tight left-to-right constraint of classical IMT systems. \citet{Marie15} proposed a system based on touch-interactions, which allowed the user to select the correct parts of a hypothesis. Extending this work, \citet{Cheng16} developed a pick-revise procedure for IMT, consisting in the selection by the user of the most critical part of a hypothesis and its correction. This pick-revise framework has been also applied to NMT systems~\citep{Hokamp17}. Related to this, \citet{Gonzalez16}, \citet{Domingo18} and \citet{Peris17a} allowed the selection of correct segments from translation hypotheses, which must be remain fixed along the IMT 
process. 

It is worth noting a major difference between these last two works and the one developed by \citet{Cheng16}: while \citet{Gonzalez16} and \citet{Peris17a} demand perfect translations for a given sentence (as in a full post-editing setup), \citet{Cheng16} accept some translation errors, sacrificing the final quality at the expense of a minor human effort (as in light post-editing). 

\subsection{Online learning in machine translation}

Online learning is a paradigm thoroughly explored in the literature. In the field of MT, most works aimed to adapt a MT system to the user or to tailor it for a given document. The most clear example is the use of OL techniques for adjusting the weights of the log-linear model of PB-SMT. A significant number of algorithms have been applied to this task. The margin-infuse relaxed algorithm (MIRA) \citep{Crammer01} processes all samples one-by-one and it becomes especially useful when dealing with a large number features. It has been applied to PB-SMT with sparse features~\citep{Watanabe07,Chiang12,Green13}.

Another common usage of OL in MT, is to perform user or domain adaptation of a system. \citet{Martinez-Gomez12} studied several OL algorithms for adjusting the weights of a PB-SMT system during the post-editing phase. Similarly, \citet{Mathur13} introduced an additional features, which allowed to take into account corrections done by the user. Closely related to this, \citet{Denkowski14} implemented dynamic translation and language models which, together with the tuning of weights from the log-linear model, provided a reduction of the human effort required for post-editing the outputs of a system.

Beyond the tuning of the weights of the log-linear model, the re-estimation of the sub-models that conform PB-SMT systems via OL has also received attention. Many advances in this direction were achieved during the CasMaCat \citep{Alabau13} and MateCat~\citep{Federico14} projects. \citet{Lagarda15} adapted a general PB-SMT system to a specific domain, during the post-editing stage. This work was extended to the interactive framework by~\citet{Ortiz16}.

Few works studied the application of OL techniques to NMT in the post-editing scenario. Almost simultaneously, \citet{Turchi17} and \citet{Peris17b} posed a similar scenario, in which an NMT system was refined with post-edited samples in order to perform domain adaptation. Both works used online SGD in the continuous learning phase. The NMT system was significantly improved in almost every case. Additionally, \citet{Peris17b} considered alternative optimization methods, but they obtained poorer results than using traditional SGD. In this work, we extend the application of OL to INMT. Moreover, we study the effectiveness of OL for NMT in multiple and varied setups.
\section{Neural machine translation}
\label{sec:NMT}

Statistical approaches to MT \citep{Brown93} aim to find the most likely sentence $\hat{y}^{\hat{I}}_1 = \hat{y}_1,\dots,\hat{y}_{\hat{I}}$ in the target language, given a sentence $x^J_1=x_1,\dots,x_J$ in the source language:
\begin{equation}
\hat{y}^{\hat{I}}_1 = \argmax_{I,y^I_1} \Pr(y^I_1\mid x^J_1) %= \argmax_{I,y^I_1} \prod_{t=1}^{I}\Pr(y_t\mid y_1^{i-1}, x^J_1)
\label{eq:SMT}
\end{equation}

NMT systems directly model this posterior probability. Most NMT approaches follow an encoder--decoder paradigm: the encoder is a neural network which computes a compact representation of the input sentence. The decoder is another neural network that takes this representation and decodes it into a sentence in the target language. Such networks are typically recurrent neural networks with LSTM units or GRU. Nevertheless, other network architectures also fit in the encoder--decoder framework~\citep{Gehring17,Vaswani17}.
Our NMT system was inspired by~\citet{Bahdanau15}, but we took into account recommendations given by~\citet{Sennrich17} and \citet{Britz17}. 

Each source sentence $x_1, \dotsc, x_J$ is processed as a sequence of words. It is inputed to the system and projected to a continuous space by means of a word embedding matrix. The sequence of word embeddings feeds a bidirectional~\citep{Schuster97} LSTM network, which concatenates the hidden states from the forward and backward layers, producing a sequence of annotations $\h_1, \dotsc, \h_J$.

The decoder is a conditional LSTM (cLSTM) network which takes into account the sequence of annotations together with the previously generated word ($y_{t-1}$). The cLSTM unit represents a novel extension of the conditional GRU (cGRU) with attention~\citep{Sennrich17} to LSTMs.

\begin{figure}[!t]
	%	\vspace{-1em}
	\centering
	\includegraphics[width=0.95\textwidth]{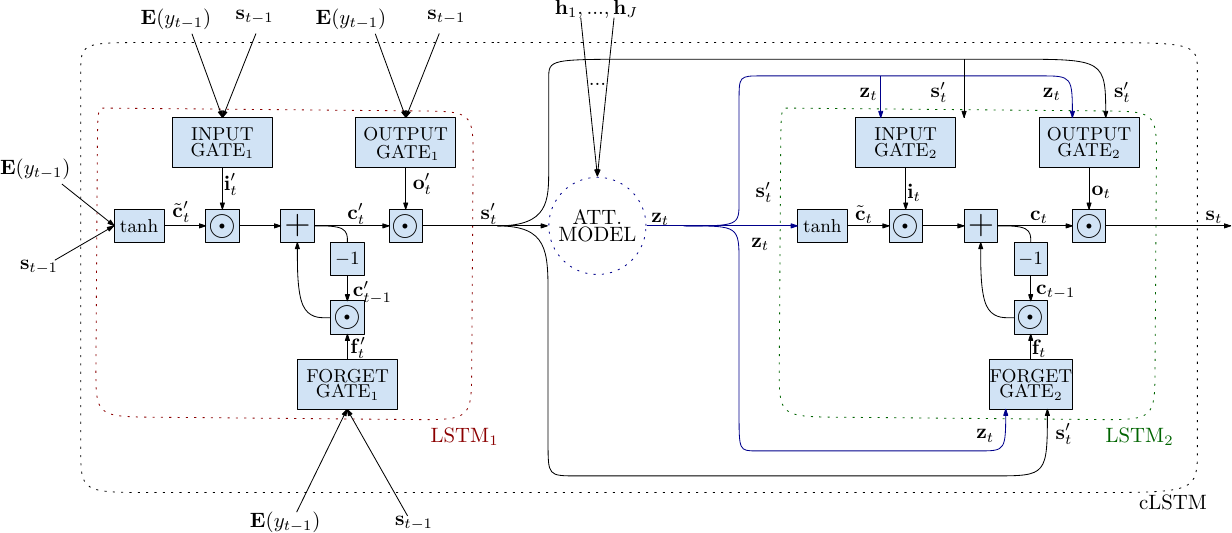}
	\caption{\label{fig:lstm}Conditional LSTM (cLSTM) cell, with attention, as used in the decoder. Dotted lines separate each one of its components: the first LSTM block (red), the attention mechanism (blue) and the second LSTM block (green). ${\E}(y_{t-1})$ is the word embedding of the previously generated word, $\s'_t$ and $\s_t$ are the hidden states of the LSTM blocks at the time-step $t$ and $\z_t$ is the context vector computed by the attention model from the sequence of annotations $\h_1, \dotsc, \h_J$ and $\s'_t$.}
\end{figure}

As cGRUs, a cLSTM unit is composed of LSTM transition blocks together with an attention mechanism. \cref{fig:lstm} shows an illustration of our cLSTM cell. 

In our case, we use two LSTM blocks. The first block, combines the word embedding of the previously generated word (${\E}(y_{t-1})$) together with the hidden state from the second LSTM block at the previous time-step ($\s_{t-1}$), obtaining the intermediate representation $\s_t'$:
\begin{equation}
\s_t'= \mathrm{LSTM}_1 ({\E}(y_{t-1}), \s_{t-1})
\label{eq:lstm1}
\end{equation}

The $\mathrm{LSTM}_1$ function is defined according to the following equations\footnote{For notation simplicity, we omit the bias terms in all expressions.}~\citep{Hochreiter97,Gers00}:
\begin{gather*}
{\s}'_t = {\mathbf o}'_t \odot {\mathbf c}'_t\\
{\mathbf c}'_t = {\mathbf f}'_t \odot {\mathbf c}'_{t-1} + {\mathbf i}'_t \odot {\tilde {\mathbf c}}'_t\\
{\tilde {\mathbf c}}'_t = \tanh({\W}'_c{\E}(y_{t-1}) + {\U}'_c{\s}_{t-1} )\\
{\mathbf f}'_t = \sigma({\W}'_f{\E}(y_{t-1}) + {\U}'_f {\s}_{t-1})\\
{\mathbf i}'_t = \sigma({\W}'_t{\E}(y_{t-1}) + {\U}'_t {\s}_{t-1})\\ 
{\mathbf o}'_t = \sigma({\W}'_o{\E}(y_{t-1}) + {\U}'_o {\s}_{t-1})
\end{gather*}
where ${\E}$ is the target text word embedding matrix and ${\E}(y_{t-1})$ denotes the word embedding of the previously generated word ($y_{t-1}$). 
${\mathbf i}'_t$, ${\mathbf o}'_t$ and ${\mathbf f}'_t$ are the input, output and forget gates, which control the information flow along the cell. ${\mathbf c}'_t$ and ${\tilde {\mathbf c}}'_t$ are the so-called cell and updated cell states, respectively. $\W'_c$, $\U'_c$, $\W'_f$, $\U'_f$, $\W'_t$, $\U'_t$, $\W'_o$ and $\U'_o$, are the trainable weight matrices. $\odot$ denotes element-wise multiplication and $\sigma$, logistic sigmoid activation function.

At each decoding time-step $t$, an attention mechanism weights every element from the sequence of annotations, according to the intermediate representations obtained in Eq. \ref{eq:lstm1}. A single-layer perceptron computes an alignment score, as in \citet{Bahdanau15}: 
\begin{equation}
e_{j t} = \w^\top \tanh(\W_a \s'_{t} + \U_a {\mathbf h}_j)
\end{equation}
where $\w$, $\W_a$ and $\U_a$ are trainable parameters. This aligner is then followed by a softmax function, for obtaining normalized weights:
\begin{equation}
\alpha_{j t} = \frac{\exp{(e_{j t})}}{\sum_k^J\exp{(e_{k t})}}. 
\end{equation}

Finally, the context vector ($\z_t$) is computed as a weighted sum of the annotations:
\begin{equation}
\z_t = \sum_{j=1}^J \alpha_{j i} {\mathbf h}_j,
\end{equation}

This context vector is the input to the second LSTM block, which also takes into account the intermediate representation $\s'_t$:
\begin{equation}
\s_t= \mathrm{LSTM}_2 (\s'_{t}, \z_{t})
\label{eq:lstm2}
\end{equation}

The $\mathrm{LSTM}_2$ transition block is similar to the $\mathrm{LSTM}_1$, but with different inputs:
\begin{gather*}
{\s}_t = {\mathbf o}_t \odot {\mathbf c}_t\\
{\mathbf c}_t = {\mathbf f}_t \odot {\mathbf c}_{t-1} + {\mathbf i}_t \odot {\tilde {\mathbf c}}_t\\
{\tilde {\mathbf c}}_t = \tanh({\W}_c\z_t + {\U}_c{\s}'_{t} )\\
{\mathbf f}_t = \sigma({\W}_f\z_t + {\U}_f {\s}'_{t})\\
{\mathbf i}_t = \sigma({\W}_t\z_t + {\U}_t {\s}'_{t})\\ 
{\mathbf o}_t = \sigma({\W}_o\z_t + {\U}_o {\s}'_{t})
\end{gather*}
where, as above, $\W_c$, $\U_c$, $\W_f$, $\U_f$, $\W_t$, $\U_t$, $\W_o$ and $\U_o$ are the trainable weight matrices; ${\mathbf i}_t$, ${\mathbf o}_t$ and ${\mathbf f}_t$ are the input, output and forget gates; and ${\mathbf c}_t$ and ${\tilde {\mathbf c}}_t$ are cell and updated cell states.

The output of the decoder ${\s}_t$ is combined together with context vector $\z_t$ and the word embedding of the previously generated word ${\E}(y_{t-1})$ in a deep output layer~\citep{Pascanu14}, to obtain an $L$-sized intermediate representation $\mathbf{t}_t$:
\begin{equation}
	\mathbf{t}_t = \tanh( \s_t \W_{t1} + \z_t \W_{t2}  + {\E}(y_{t-1}) \W_{t3} )
\end{equation}
where $\W_{t1}$, $\W_{t2}$ and $\W_{t3}$ are trainable weight matrices.

Finally, the probability of the word $y_t$ at time-step $t$ is defined as:

\begin{equation}
p(y_t \mid {y_1^{t-1}},  x)  = \mathbf{\bar{y}}_t^\top \text{softmax}(\V	\mathbf{t}_t)
\label{eq:word-prob}
\end{equation}
where $\V \in \mathbb{R}^{\vert V_y \vert \times L}$ is a weight matrix, $\vert V_y \vert$ is the size of the target language vocabulary and $\mathbf{\bar{y}}_t \in [0, 1]^{\vert V_y \vert}$ is the one-hot codification of word $y_t$.

\subsection{Training and decoding}
\label{sec:nmt-training}
 All model parameters $\theta$ (the weight matrices of the neural networks) are jointly estimated on a parallel corpus ${\cal S} = \{(x^{(s)}, y^{(s)})\}_{s=1}^S$, consisting of $S$ sentence pairs. The training objective is to minimize a loss function $\ell_\theta$, typically the minus log-likelihood, over $\cal S$:
 \begin{equation}
 \begin{gathered}
 \label{eq:training-obj}
 \widehat{\theta} = \argmin_{\theta} \ell_\theta ({\cal S}) = \\
 = \argmin_{\theta}\sum_{s=1}^S \sum_{t=1}^{I_s} 
 - \log(p_\theta(y_t^{(s)} \mid {y_1^{t-1}}^{(s)},  x^{(s)}))  
 \end{gathered}
 \end{equation}
 where $I_s$ is the length of the $s$-th target sentence and ${y_1^{t-1}}^{(s)}$ denotes the $s$-th target sentence up to the position $t-1$.
  
A beam search method is used at decoding time, in order to find the most likely translation \citep{Bahdanau15,Sutskever14}.

\section{Interactive neural machine translation}
\label{sec:IMT}
\begin{wrapfigure}{r}{0.3\textwidth}
	\centering
	\includegraphics[width=0.28\textwidth]{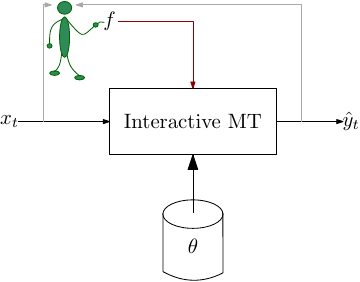}
	\caption{\label{fig:imt} IMT framework. The MT system generates a translation hypothesis $\hat{y}_t$ of the source sentence $x_t$. The user reviews this hypothesis, and introduces a feedback signal $f$. In the next iteration, the system will consider $f$ for generating a new (and hopefully better) hypothesis.}
\end{wrapfigure}

In the general framework of IMT~\citep{Barrachina09}, a source sentence $x_1^J$ is inputted to the system, which produces a translation hypothesis $\hat{y}_1^{\hat {I}}$. Next, a human agent reviews the hypothesis and provides a feedback signal $f$. The IMT system produces a new translation hypothesis, considering this user feedback. It is expected that this new hypothesis is better than the previous one, as the system has more information. Then, a new iteration starts. This iterative procedure continues until the user accepts the hypothesis provided by the system. \cref{fig:imt} shows an illustration of the IMT framework.

The feedback signal can range from a simple word correction to complex, indeterministic ones, such as an eye-gaze tracking, combined with a handwritten correction, as in \citet{Alabau13}. The features of the signal and its meaning determine the nature and behavior of the IMT system. In this work, we use keyboard and mouse to introduce feedback to the INMT system.

From a probabilistic point of view, the inclusion of the feedback signal affects~\cref{eq:SMT}, which now is also conditioned to $f$:
\begin{equation}
\hat{y}^{\hat{I}}_1 = \argmax_{I,y^I_1} \Pr(y^I_1\mid x^J_1, f) 
\label{eq:IMT}
\end{equation}

In this work, we refine the prefix-based interactive approach presented by~\citet{Peris17a}, who extended this IMT protocol to NMT. In this case, the feedback signal provided by the user contained the correction of the leftmost wrong word from the translation hypothesis, located at position $t'$. With this, the user inherently validated a translation prefix (up to $t'$). Hence, $f$ can be seen as a validated prefix: $f=\hat{y}_1^{t'}$. Next, the system provided an alternative hypothesis, which contained the validated prefix together with an alternative suffix. Therefore, \cref{eq:IMT} was reformulated as:
\begin{equation}
\hat{y}^{\hat{I}}_1 = \argmax_{I,y^I_{t'+1}} \Pr(y^I_{t'+1}\mid x^J_1, \hat{y}_1^{t'}) 
\label{eq:INMT}
\end{equation}
which implies a search over the space of translations, but constrained by the validated prefix $\hat{y}_1^{t'}$.

The application of this expression to NMT implies a modification of the search method, for taking into account the validated prefix $\hat{y}_1^{t'}$. Therefore, the probability of a word (Eq. \ref{eq:word-prob}) is now computed as:
\begin{equation}
\label{eq:inmt-prefix}
p(y_{t} \mid y_1^{t-1}, x_1^J, \hat{y}_1^{t'}) = 
\begin{cases} \delta(y_{t}, \hat{y}_{t'}), & \mbox{if } {t \leq t'} \\
\bar\y_{t}^\top \text{softmax}(\V	\mathbf{t}_t) 
& \mbox{otherwise }\end{cases}
\end{equation}
where $\delta(\cdot, \cdot)$ is the Kronecker delta. As in \cref{eq:word-prob}, $\mathbf{\bar{y}}_t $ is the one-hot codification of word $y_t$ and $\text{softmax}(\V\mathbf{t}_t)$ is the output layer of the NMT system.

\subsection{Vocabulary masking for character-level INMT}
\label{sec:INMT-mask}
The INMT system developed by~\citet{Peris17a} performed all interactions at a word level. Nevertheless, it is interesting to enables the user to interact with the system at a character level. This makes possible a major granularity and a more natural interaction with the system. Most of the existing IMT tools already accept character-level interactions (e.g. CasMaCat). 

In the field of NMT, to perform translations at character-level is a promising research direction \citep{Costa16,Chung16}. Unfortunately, the prohibitive decoding times of character-level NMT~\citep{Lee16} prevent its direct usage in an interactive setup.

In this work, we propose a simple, yet effective way for interacting with the INMT system at character level. The feedback signal provided by the user is at character-level. For the sake of simplicity, we stick to prefix-based interaction. That is, the user will correct the hypotheses from the left to the right. Therefore, the user inputs the leftmost wrong character of a hypothesis. Nevertheless, the same method is extensible to other interactive protocols (e.g. segment-based interaction \citep{Peris17a}).

\begin{wrapfigure}{r}{0.3\textwidth}
	%	\vspace{-1em}
	\centering
	\includegraphics[width=0.28\textwidth]{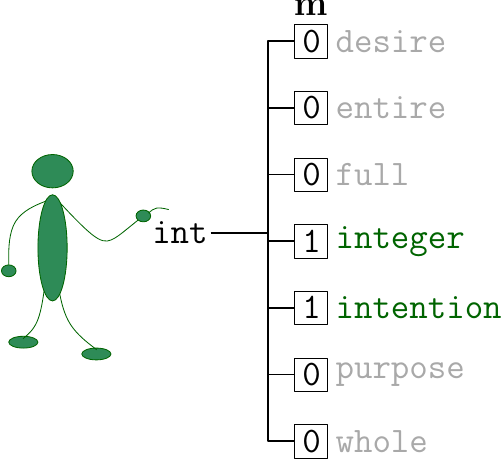}
	\caption{\label{fig:prefix_filtering} Constraining the vocabulary for character-level interaction.  For this example, we assume a vocabulary of 7 words. The user introduces a valid prefix consisting one or more characters (\texttt{int}). For predicting the next word, the system computes a compatibility mask $\m$, which filters those words that are incompatible with the given prefix (in gray). In this case, the compatible words (in green) are \texttt{integer} and \texttt{intention}.}
\end{wrapfigure}

As before, the system must produce a hypothesis compatible with the user feedback. In this case, the user introduced a character correction in the $u$-th position of the $t'$-th word. Therefore, the validated prefix are all words up to position $t'-1$ together with the validated part of the $t'$-th word:
\begin{displaymath}
f = \hat{y}_0^{t'-1},\hat{y}_{{t'}_{1}^{u}}
\end{displaymath}
where $\hat{y}_0^{t'-1}$ is the sequence of validated words, up to word in position $t'-1$, $\hat{y}_{{t'}_{1}^{u}}$ is the correct part of the word $\hat{y}_{t'}$ together with the corrected character position $u$.

When processing this signal, we may need to generate a word constrained by the prefix $\hat{y}_{{t'}_{1}^{u}}$ or not.
The latter case is handled by the classical NMT system without modifications.

For tackling the first case, we create a mask $\m_u$ of the target vocabulary according to the user prefix~$\hat{y}_{{t'}_{1}^{u}}$. Therefore, $\m_u \in  [0, 1]^{\vert V_y \vert}$ is a vocabulary-sized binary vector, in which each position is set to $1$ if the corresponding word in the vocabulary is compatible with the user prefix and to $0$ otherwise. If there are no compatible words with the validated prefix, we apply forced decoding to this prefix and continue the process with the unconstrained vocabulary. \cref{fig:prefix_filtering} shows an example of this masking strategy. 

The word probability expression (\cref{eq:inmt-prefix}) is then computed as:

\begin{equation}
\label{eq:inmt-prefix-char}
p(y_{t} \mid y_0^{t-1}, x_1^J, \hat{y}_1^{t'-1}, \hat{y}_{{t'}_{1}^{u}}) = 
\begin{cases} \delta(y_{t}, \hat{y}_{t'}), & \mbox{if } {t < t'} \\
\mathbf{m}_u^\top \bar\y_{t}^\top \text{softmax}(\V	\mathbf{t}_t), & \mbox{if } {t = t'} \\
\bar\y_{t}^\top \text{softmax}(\V	\mathbf{t}_t) 
& \mbox{otherwise }\end{cases}
\end{equation}

With this strategy, we get the benefits of character-level interaction while maintaining the decoding speed of (sub)word-level NMT. 
Moreover, since we keep the probabilities of each compatible word, is straightforward to implement additional features to the system, such as word completion.

It is also remarkable that this vocabulary masking strategy can help the system to disambiguate words. For instance, by looking at \cref{fig:prefix_filtering}, if we filter the vocabulary, the system must choose between \texttt{integer} and \texttt{intention}. This reduces the possible ambiguity with other vocabulary words, such as \texttt{entire}, \texttt{full} or \texttt{whole}. 

\section{Online learning in NMT post-editing}
\label{sec:OL}
\begin{figure*}[!h]
	\centering	
	\begin{subfigure}[t]{0.35\textwidth}
		\centering
		\includegraphics{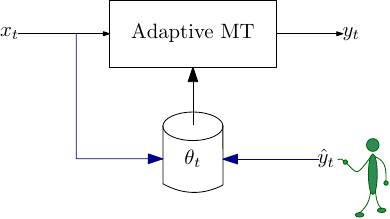}
		\caption{\label{fig:onlineMT} Machine translation post-editing.}
	\end{subfigure}%
	\qquad	\qquad	\qquad	\qquad
	\begin{subfigure}[t]{0.35\textwidth}
		\centering
		\includegraphics{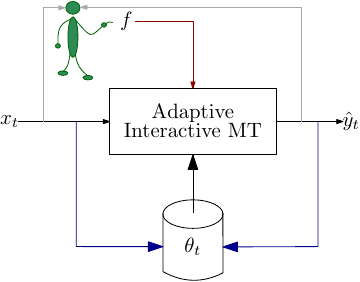}
		\caption{\label{fig:onlineIMT} Interactive machine translation.}
	\end{subfigure}
	\caption{\label{fig:OL-MT}Online learning in MT post-editing and IMT. The system translates a source sentence $x_t$, producing a hypothesis $y_t$. This hypothesis is corrected. The corrected hypothesis $\hat{y}_t$ is used, together with $x_t$ by the system to modify its models (parametrized by $\theta_t$). The difference between classical post-editing and IMT is the way that the hypothesis is corrected.}
\end{figure*}

Under an online learning paradigm, data comes available sequentially and models are updated incrementally. The typical post-editing or IMT scenario perfectly matches with these stages:

\begin{enumerate}
\item A new source sentence $x_t$ comes to the system.
\item The system produces a translation hypothesis $y_t$.
\item A human agent revises this hypothesis and corrects the errors made by the system or interactively translates the source sentence. This generates a correct translation $\hat{y}_t$.
\item The system uses the corrected sample to adapt its models. 
\end{enumerate}

Following this procedure, we can build adaptive MT systems, able to take into account corrections made by the humans. \cref{fig:OL-MT} illustrates the adaptive MT framework, either with IMT or classical post-editing. 

Adaptive NMT systems have interesting applications in the translation industry. The NMT technology has good results if the training corpora are large enough~\citep{Koehn17}. But to acquire high amounts of parallel corpora is an expensive process. Moreover, for building a NMT system for a given domain, we need data from this domain, which can be difficult to obtain. 
A common approach is to train a system on a large, general corpus and fine-tune it with in-domain data. Nevertheless, this is may also be infeasible; for instance, if the domain is unknown.

Therefore, continuous learning from MT post-edits or IMT constitute techniques that can be exploited for adapting a MT system to different domains or styles. They are orthogonal to other adaptation techniques, like fine-tuning with in-domain data. In this work, we apply OL to NMT systems for performing domain adaptation of general NMT systems. Moreover, we also study the power of OL for refining an NMT system already trained with in-domain data.

The most common training procedure of neural networks is SGD (\cref{sec:nmt-training}), which can directly be applied in an online way. Therefore, the application of OL to NMT becomes natural. Online adaptation of NMT systems can be performed with the same optimizers than used during (mini-batch) training, but sample-to-sample.

For a training sample $(x_t, \hat{y}_t)$, SGD updates the parameters following the direction of the gradient of the objective function $\ell$ (\cref{eq:training-obj}) with respect to the weights $\theta_t$:
\begin{equation}
\label{eq:grad}
\Delta \theta_t = - \rho~ \nabla \ell_{\theta_t}(x_t,\hat{y}_t)
\end{equation}
where $\nabla \ell_{\theta_t}$ is the gradient of $\ell$ with respect to $\theta_t$ and $\rho$ is a learning rate that controls the step size.

This update rule relies on a careful choice of $\rho$. A significant effort has been spent in the literature trying to minimize the critical importance of the learning rate choice. Therefore, the so-called adaptive SGD algorithms try to overcome this dependence by dynamically computing the learning rate. 

Among the most common adaptive optimizers, we find AdaGrad~\citep{Duchi11}, which updates the weights according to the sum of the squares of the past gradients; Adam~\citep{Kingma14} which computes decaying averages for the past gradients and past squared gradients or Adadelta~\citep{Zeiler12}, which updates the parameters according to the root mean square (RMS) of the gradients and corrects these updates according to the RMS of the previous update.

Nevertheless, in our scenario of online learning in NMT, the usage of adaptive SGD algorithms does not completely alleviate the learning rate tuning. We found (\cref{sec:results-1}) that a correct choice of the learning rate is extremely important, even for adaptive SGD optimizers, to make them properly work.

\section{Experimental framework}
\label{sec:ExperimentalFramework}

We conducted an exhaustive experimentation in order to assess the effectiveness of our proposals. This section details the experimental setup, evaluation metrics, simulation procedure, corpora and the translation systems.

\subsection{Corpora}
\label{sec:corpora}
We evaluated our models in 5 tasks of different complexity and nature. For all corpora, we performed the translation from English to German and from English to French. First, we used two corpora extensively used in the IMT literature: The XRCE and EU corpora~\citep{Barrachina09}. The former consists of printer manuals from Xerox printers and the latter is a collection of proceedings collected from the Bulletin of the European Union. For these tasks, we used the default data splits. We also tested our proposals with data from WMT'17~\citep{Bojar17}, using the Europarl and the UFAL corpora. Europarl~\citep{Koehn05} collects the proceedings from the European Parliament. For the sake of comparison with \citet{Ortiz16}, we used the \texttt{newstest2012} and the \texttt{newstest2013} as development and test partitions respectively. The UFAL medical corpus\footnote{\url{https://ufal.mff.cuni.cz/ufal_medical_corpus}} contains data crawled from several medical collections, collected during the European project \textit{Health in my Language} \citep{Bojar17c}. We used the development and test data from the Khresmoi project \citep{Libovicky16}. Finally, the TED task \citep{Cettolo12} refers to the translation of TED talks and it has also been used in IMT and online learning works. We used the standard \texttt{dev2010} and \texttt{tst2010} partitions for development and test, respectively.

All corpora were tokenized using the Moses scripts. For training the NMT systems, we applied joint byte-pair-encoding (BPE)~\citep{Sennrich16} to all corpora, using $32,000$ merge operations. Table~\ref{table:corpora} shows the main figures of each corpus, after tokenization. For the test set, we show additional metrics, aiming to obtain an estimation of the potential efficacy of the OL process: 

\begin{description}
	\item [Repetition rate (RR)]\citep{Bertoldi13}: measures the repetitiveness of a document. It is computed as the rate of non-singleton $n$-grams (with $n$ from 1 to 4). The rates are obtained through a sliding window of $1,000$ words and geometrically averaged.
	
	\item [Restricted repetition rate (RRR)]\citep{Ortiz16}: RR is unable to capture whether a specific $n$-gram from the text to translate was in the document used for training the models. Therefore, it may be insufficient to properly estimate the potential of OL. The RRR aims to overcome this issue by computing the RR on those $n$-grams from the test not contained in the training data.
	
	\item [Unseen $n$-gram fraction (UNF)]\citep{Ortiz16}: ratio of unseen $n$-grams in the test document. As in the RR, we consider $n$-grams from order $1$ up to $4$.
\end{description}

The effectiveness of OL can be estimated beforehand by paying attention to these three metrcs metrics~\citep{Ortiz16}. OL will likely be more effective for documents tasks with high values of RR, RRR and UNF.

\begin{table}[th]
	\centering
	\caption{\label{table:corpora}Main figures of the XRCE, EU, UFAL, Europarl and TED corpora. $\vert S \vert$, $\vert T \vert$ and $\vert V \vert$, RR$~[\%]$ account for number of sentences, number of tokens, vocabulary size, repetition rate, restricted repetition rate unseen $n$-gram fraction, respectively. RR, RRR and UNF were computed after the BPE process. k and M stand for thousands and millions.}
	{  
		\footnotesize
		\begin{tabular}{llrrrrrrrrrrrr}
			\toprule
			& & \multicolumn{3}{c}{Training}
			& \multicolumn{3}{c}{Development}
			& \multicolumn{6}{c}{Test} \\ \cmidrule(lr){3-5}\cmidrule(lr){6-8}  \cmidrule(lr){9-14}
			&  & $\vert S \vert$ &$\vert T \vert$ & $\vert V \vert$ & $\vert S \vert$ & $\vert T \vert$ &$\vert V \vert$ &$\vert S \vert$ &$\vert T \vert$ & $\vert V \vert$ & RR$~[\%]$ & RRR$~[\%]$ &UNF$~[\%]$ \\
			\midrule
			\multirow{4}{*}{XRCE}  &
			De  & \multirow{2}{*}{$50k$}   & $531k$ & $23k$ &  \multirow{2}{*}{$964$} & $10k$ & $2k$ &  \multirow{2}{*}{$995$}& $12k$ & $2k$ & $23.6$ & $17.4$ & $25.7$ \\
			&
			En  &  & $587k$ & $11k$ &  & $11k$ & $1k$  & & $12k$ & $2k$& $28.0$& $16.0$ & $9.1$ \\
			\cmidrule(lr){2-14}
			& Fr  & \multirow{2}{*}{$52k$}   & $676k$ & $16k$ & \multirow{2}{*}{$994$} & $12k$ & $2k$ &\multirow{2}{*}{$984$}&  $12k$ & $2k$  & $27.9$& $18.3$ & $31.5$ \\
			&
			En  &  & $615k$ & $15k$ &  & $11k$ & $2k$  &  & $11k$ & $2k$ & $26.9$& $18.3$ & $12.6$ \\
			\midrule
			\multirow{4}{*}{EU}  &
			De  & \multirow{2}{*}{$222k$} & $5.4M$   & $109k$ & \multirow{2}{*}{$400$} & $10k$ & $3k$ &\multirow{2}{*}{$800$}&  $19k$  & $5k$& $9.3$& $0.0$ & $3.0$ \\
			&En  &  & $5.7M$ & $42k$ & & $10k$  & $2k $ && $20k$  & $4k$ & $12.3$& $0.7$ & $3.5$ \\
			\cmidrule(lr){2-14}
			& Fr  & \multirow{2}{*}{$215k$} & $6.2M$  & $58k$ & \multirow{2}{*}{$400$} &  $12k$ & $3k$ &  \multirow{2}{*}{$800$}  & $24k$ & $4k$ & $14.0$&$0.0$ & $0.0$ \\  
			& En  &  & $5.2M$ & $50k$ & & $10k$  & $3k$ && $20k$  & $4k$& $12.3$&   $0.0$ & $2.4$ \\
			
			\midrule
			 \multirow{4}{*}{UFAL} &
			De  & \multirow{2}{*}{$3.0M$}  & $109M$ & $1.6M$  &  \multirow{2}{*}{$500$} & $10k$ &  $3k$  &\multirow{2}{*}{$1,000$} & $19k$ & $3k$ & $10.7$& $3.6$ & $4.2$ \\
			&
			En  &  & $116M$ & $671k$ &   & $10k$ & $3k$ &  & $21k$ &  $4k$& $13.4$& $0.0$ & $0.0$  \\
			\cmidrule(lr){2-14}                                                                                            
			& Fr  & \multirow{2}{*}{$2.8M$}  & $125M$ & $647k$  &  \multirow{2}{*}{$500$} & $12k $&  $3k$  &\multirow{2}{*}{$1,000$} & $26k$ & $5k$& $16.6$& $5.9$ & $3.3$  \\
			&
			En  &  & $109M$ & $655k$ &   & $10k$ & $3k$ &  & $21k$ & $4k$& $13.1$& $0.0$ & $4.3$  \\
			\midrule
			
			\multirow{4}{*}{Europarl}  &
			
			De  & \multirow{2}{*}{$1.9M$} & $50M$  & $393k$ & \multirow{2}{*}{$3,003$}  & $73k$ & $14k$  &\multirow{2}{*}{$3,000$} & $65k$  & $12k$ & $13.4$& $8.9$ & $18.4$ \\
			&
			En  &   & $53M$ & $123k$ &   & $73k$ & $10k$ &  & $63k$ & $10k$ &$15.0$& $9.4$ & $15.4$ \\
			\cmidrule(lr){2-14}                            
			& Fr  & \multirow{2}{*}{$2.0M$}  & $61M$   & $153k$ & \multirow{2}{*}{$3,003$}  & $82k$  &$11k$ &\multirow{2}{*}{ $3,000$} & $74k$  & $11k$& $12.6$ & $8.1$ & $7.5$ \\
			&
			En  &  & $56M$  & $134k$ &  & $71k$  & $10k$  &  & $65k$ & $10k$ &$14.2$& $6.6$ & $7.9$  \\
			\midrule
			
			\multirow{4}{*}{TED}  &
			
			De  & \multirow{2}{*}{$133k$}   & $2.4M$ & $102k$ & \multirow{2}{*}{$883$}  & $19k$ & $4k$ &\multirow{2}{*}{$1,565$} & $30k$  & $5k$& $11.2$& $4.7$ & $3.7$  \\
			&
			En  &   & $2.6M$ & $50k$ &   &  $20k$ & $3k$  &  &$32k$ &  $4k$& $14.6$& $7.0$ & $4.8$ \\
			\cmidrule(lr){2-14}                            
			& Fr  & \multirow{2}{*}{$107k$}   & $2.2M$ & $58k$ & \multirow{2}{*}{$934$} & $20k$ & $4k$  & \multirow{2}{*}{$1,664$}&  $34k$ & $5k$ & $14.8$ & $4.3$ & $7.2$ \\
			&
			En  &  & $2.1M$ & $47k$ &  & $20k$ & $3k$ & & $32k$ & $4k$& $14.8$& $5.2$ & $6.3$ \\                         
			\bottomrule
			
		\end{tabular}
	}
	
\end{table}

\subsection{Evaluation}
\label{sec:evaluation}
The ultimate goal of this work is to reduce the human effort required in the MT supervision process. Therefore, we must assess either the translation quality of the MT system and, more importantly, this human effort. We evaluated both, translation quality and human effort.

\subsubsection{Translation quality metrics}

Quality assessment of MT is an open problem. The main goal of an automatic metric is to achieve a high correlation with human judgment of the quality of a translation~\citep{Bojar17b}. For doing this, the MT translation hypotheses are usually compared to a reference translation. We evaluated the quality of the MT systems according to two common metrics:

\begin{description}
	\item [Translation Edit Rate (TER)]\citep{Snover06}: minimum number of edit operations required for converting the translation hypothesis into the reference, divided by the number of reference words. The edit operations considered are insertion, deletion, replacement and word sequences shifting. The minimum number of operations is obtain by means of dynamic programming. Following \citet{Zaidan10}, we use the TER as a representation of human-targeted TER, considering the reference sentences as human post-edited versions of the MT hypotheses. This gives us a broad approximation of the effort required for post-editing a translation hypothesis.
	
	\item [BiLingual Evaluation Understudy (BLEU)]\citep{Papineni02}: geometric mean of $n$-gram matchings between hypothesis and reference, modified by a brevity penalty. We use BLEU for having additional insights about the complexity of the task and the performance of the systems in terms of translation quality.
\end{description}

\subsubsection{Human effort metrics}

Human agents are involved during the IMT process, therefore, we must estimate the effort spent. We are aware that the correct way of doing this is to conduct an experimentation with real users. Nevertheless, prior to this, we need to automatically assess the effort required in the IMT scenario, in order to construct and develop efficient IMT systems. Therefore, we assumed that the reference sentences are the translations desired by the user. We estimated the human effort required according to:

\begin{description}
	
\item [Keystroke and mouse-action ratio (KSMR)]\citep{Barrachina09}: number of keystrokes plus number of mouse actions required in the IMT process, divided by the number of characters of the reference. Therefore, the lower KSMR, the better. If the user is correcting contiguous characters, no mouse action is needed. An additional mouse action that accounts for the acceptation of a hypothesis is added to each sentence.
\end{description}

\subsection{Simulation protocol}

Due to the aforementioned reasons, we evaluated our proposals with simulated users, assuming that the reference sentences are the translations that the users have in mind for each source sentence. 

In the case of IMT, since we use the prefix-based protocol, we searched for the leftmost wrong character of a translation hypothesis, comparing it with the reference. Next, we introduced the correct character, inputting this feedback signal to the system. With this, the IMT engine produced a new hypothesis that takes into account the user feedback (\cref{eq:inmt-prefix-char}). This process continued until translation hypothesis and reference matched. The user would then validate the hypothesis and finishing the process. 

In the case of the post-editing scenario, we translated a source sentence. Next, the user would edit the translation hypothesis, producing the desired translation. We directly used the reference sentences as those desired translations. 

Finally, we apply OL (as described in \cref{sec:OL}) using these edited sentences, performing the adaptation sample to sample. That is, before starting the translation of the next sentence, the models are updated according the previous sample in both the post-editing and IMT scenarios.

\subsection{Machine translation systems}
\label{sec:training_systems}
We built our NMT systems using NMT-Keras\footnote{Implementations of all models, algorithms and search methods are publicly available at \url{https://github.com/lvapeab/nmt-keras/tree/interactive_NMT}.} \citep{Peris18b}. Our systems can be implemented using either Tensorflow~\citep{Abadi16} or Theano~\citep{Theano16}. In our experiments, used the latter framework. Taking advantage of the extensive experimentation conducted by \citet{Britz17}, we set the dimensions of encoder, decoder, attention model and word embedding to $512$. Due to hardware restrictions, we used a single layer for encoder and decoder. The encoder was a LSTM network, while the decoder was made of cLSTM units (see \cref{sec:NMT}).

The learning algorithm for all base NMT systems was Adam \citep{Kingma14}, with a learning rate of $0.0002$, as in \citet{Wu16}. We clipped the $L_2$ norm of the gradients to $5$, in order to avoid the exploding gradient effect \citep{Pascanu13}. The batch size was set to $50$ and the beam size to $6$. Since we are in an interactive scenario, we discarded the usage of model ensembles at this point of the study, fostering decoding and retraining speed. We trained all models on a single GeForce GTX 1080 GPU. 

As regularization strategies, we applied Gaussian noise to the weights during training~\citep{Graves11} and batch normalizing transform \citep{Ioffe15}. We early stopped the training according to the BLEU of the development set. BLEU was checked each $5,000$ updates and the patience was set to $20$.

For the sake of comparison, we also include results of PB-SMT. We estimated PB models with the standard setup of Moses~\citep{Koehn07}: 5-gram language model with modified Kneser-Ney smoothing~\citep{Kneser95}, built with SRILM ~\citep{Stolcke02}. The phrase table was obtained using symmetrised  word
alignments, computed by GIZA++~\citep{Och02}. The model also include  distortion and reordering models. The weights of the log-linear model was optimized using Minimum Error Rate Training (MERT)~\citep{Och03}.

\section{Results and discussion}
\label{sec:Results}
In this section, we show and discuss the results obtained. For all result in this paper, we computed confidence intervals (at a 95\% level). Confidence intervals were obtained by means of pairwise bootstrap resampling~\citep{Koehn04}.

We posed three different experimental conditions, varying the amount and type of training data available. In each case, we compared the performance of adaptive versus static systems, in translation post-editing and IMT. The evaluation was always carried out on the test set of each task. Continuous learning techniques were also applied exclusively on the same test data. Therefore, the three cases we studied  are:

\begin{enumerate}
\item Exclusive availability of in-domain data.
\item Lack of in-domain data.
\item Availability of in-domain and out-of-domain data.
\end{enumerate}

\subsection{Scenario \#1: Availability of in-domain data}
\label{sec:results-1}

We first assume that we have enough in-domain data for training a system. Therefore, we followed the traditional pipeline in MT: we trained translation systems using corpora from a given domain and translated documents from the same domain. In this case, online learning techniques aim to refine each system to the test documents.

For the sake of comparison, we first evaluated PB-SMT and NMT systems for performing classical translation post-editing. We mostly focused on TER, using it as an estimation of the human effort required for post-editing the output of a MT system (see \cref{sec:evaluation}). The results of PB-SMT and NMT systems are shown in \cref{table:results-mt-quality}. 

\begin{table}[!h]
	\caption{\label{table:results-mt-quality}Results of translation quality for all tasks in terms of TER [\%] and BLEU [\%]. We compare PB-SMT and NMT systems. Results that are statistically significantly better for each task and metric are boldfaced. }
	\centering
	\small
	\begin{tabular}{llllll}
		\toprule
		& & \multicolumn{2}{c}{TER [\%]} 
		& \multicolumn{2}{c}{ BLEU [\%]}  
		\\ \cmidrule(lr){3-4}\cmidrule(lr){5-6} 
		& &PB-SMT& NMT& PB-SMT& NMT\\
		\midrule
		\multicolumn{1}{ c }{\multirow{2}{*}{XRCE} } &
		En$\rightarrow$De  & $62.5 \pm 1.0$ & $64.1 \pm 1.1$ & $24.7 \pm 0.9$ & $24.4 \pm 1.1$\\ 
		\multicolumn{1}{ c  }{}                        &
		En$\rightarrow$Fr   & $\mathbf{49.9 \pm 1.0}$ & $54.0 \pm 1.1$ &  $ \mathbf{37.1 \pm 0.9}$  & $ 34.7 \pm 1.2$ \\ 
		\midrule
		\multicolumn{1}{ c }{\multirow{2}{*}{EU} } &
		En$\rightarrow$De   & $54.1 \pm 1.0$ & $54.6 \pm 1.0$  & $35.3 \pm 1.1$ & $35.8 \pm 1.1$  \\ 
		\multicolumn{1}{ c  }{}                        &
		En$\rightarrow$Fr   &  $41.5 \pm 0.8$ & 	$\mathbf{39.3 \pm 0.8}$ & $47.1\pm 0.9$ & $\mathbf{50.0 \pm 0.9}$  \\ 
		\midrule
		\multicolumn{1}{ c }{\multirow{2}{*}{UFAL} } &
		En$\rightarrow$De  & $62.4 \pm 0.5$  & $\mathbf{56.5 \pm 0.6}$ &  $17.3 \pm 0.5$  & $\mathbf{24.2 \pm 0.5}$ \\ 
		\multicolumn{1}{ c  }{}                        &
		En$\rightarrow$Fr  & $47.5 \pm 0.5$ &$\mathbf{46.4 \pm 0.6}$& $35.0 \pm 0.5$ &  $\mathbf{37.2 \pm 0.6}$  \\ 	
		\midrule
		\multicolumn{1}{ c }{\multirow{2}{*}{Europarl} } &
		
		En$\rightarrow$De  & $\mathbf{62.2 \pm 0.3}$ & $63.1 \pm 0.4$ & $ 19.5 \pm 0.3$ & $20.0 \pm 0.3$ \\ 					
		\multicolumn{1}{ c  }{}                        &				
		En$\rightarrow$Fr  & $56.1 \pm 0.3$ & $\mathbf{55.0 \pm 0.3}$ & $26.5 \pm 0.3$ & $\mathbf{27.8 \pm 0.3}$  \\ 
		
		\midrule
		\multicolumn{1}{ c }{\multirow{2}{*}{TED} } &
		En$\rightarrow$De  & $58.4 \pm 0.5$ & $\mathbf{55.5 \pm 0.6}$ & $20.3 \pm 0.4$ & $\mathbf{24.5 \pm 0.5}$  \\ 
		\multicolumn{1}{ c  }{}                        &
		En$\rightarrow$Fr  & $51.4 \pm 0.5$ & $51.5 \pm 0.5$ & $29.9 \pm 0.5$ & $\mathbf{32.1 \pm 0.5}$ \\ 
		\bottomrule
	\end{tabular}
\end{table}

In general, the NMT approach performed slightly better than PB-SMT systems, achieving significant TER improvements for five language pairs. Only in two cases PB-SMT systems significantly outperformed NMT. In the rest of cases, differences were non-significant. Hence, human effort required for post-editing the NMT outputs was usually lower than the required for post-editing PB-SMT systems, although such differences were small.
BLEU behaviored similarly to TER: in 6 cases, NMT obtained significantly better translation hypotheses than PB-SMT. Only in one language pair, the PB-SMT system was able to significantly outperform NMT.

\label{sec:scenario1}

\begin{table}[h]

	\caption{\label{table:results-task1-literature} Effort required by interactive NMT systems (INMT) compared to the state of the art, in terms of KSMR~[\%]. $\nabla$ represents absolute decrements in KSMR percentage. Results that are statistically significantly better for each task and metric are boldfaced. $^\dagger$ refers to \citet{Ortiz16}, $^\ddagger$ to \citet{Barrachina09} and $-$ indicates missing results in the literature.}  

	\centering
	\small
	\begin{tabular}{llllr}
		\toprule
		& & \multicolumn{3}{c}{KSMR [\%]} 
		\\ \cmidrule(lr){3-5}
		& &INMT& Best in literature & $\nabla$ \\
		\midrule
		\multicolumn{1}{ c }{\multirow{2}{*}{XRCE} } &
		En$\rightarrow$De  & $\mathbf{ 32.2 \pm 0.7}$ &$40.1  \pm 1.2$$^\dagger$ & $7.9$ \\
		\multicolumn{1}{ c  }{}                        &
		En$\rightarrow$Fr & $ \mathbf{27.5 \pm 0.6}$ & $ 35.8 \pm 1.3$$^\dagger$ & $8.3$ \\
		\midrule
		\multicolumn{1}{ c }{\multirow{2}{*}{EU} } &
		En$\rightarrow$De & $\mathbf{19.8 \pm 0.5}$ &  $30.5 \pm 1.1$$^\ddagger$ & $10.7$ \\
		\multicolumn{1}{ c  }{}                        &
		En$\rightarrow$Fr   & $\mathbf{16.3 \pm 0.4}$ &$ 25.5 \pm 1.1$$^\ddagger$ & $9.2$ \\
		\midrule
		\multicolumn{1}{ c }{\multirow{2}{*}{UFAL} } &
		En$\rightarrow$De    & $ 22.5 \pm 0.3$ &  $-$  & $-$ \\
		\multicolumn{1}{ c  }{}                        &
		En$\rightarrow$Fr & $ 19.7 \pm 0.3$ & $-$  & $-$ \\
		\midrule
		\multicolumn{1}{ c }{\multirow{2}{*}{Europarl} } &
		En$\rightarrow$De    & $ \mathbf{32.3 \pm 0.2}$ & $ 49.2 \pm  0.4$$^\dagger$ & $16.9$ \\
		\multicolumn{1}{ c  }{}                        &
		En$\rightarrow$Fr & $\mathbf{ 29.8 \pm 0.2} $ &$44.4 \pm 0.5 $$^\dagger$ & $14.6$ \\
		\midrule
		\multicolumn{1}{ c }{\multirow{2}{*}{TED} } &
		En$\rightarrow$De   & $28.0 \pm 0.3$ & $-$ & $-$ \\
		\multicolumn{1}{ c  }{}                        &
		En$\rightarrow$Fr& $26.8 \pm 0.3$ & $-$ & $-$ \\
		\bottomrule
	\end{tabular}
\end{table}

Next, we move to interactive machine translation. %We required perfect translations after the IMT process, ensuring a fair comparison between post-editing and IMT.
\cref{table:results-task1-literature} shows the performance in \%~KSMR of the INMT systems. We also compare these results with the best results obtained in the literature for each task. All IMT systems from the literature were PB-SMT. 

In terms of effort, INMT systems were substantially better than all state-of-the-art systems. In all cases, the effort was greatly reduced: from $7.9$ to $16.9$ points in KSMR percentage. This approximately accounts from a $20\%$ to a $36\%$ relative improvement. 

This improvement of INMT systems over classical PB-SMT systems was also reported by \citet{Knowles16} and \citet{Peris17a}, who found that the INMT technology reacted much better to user feedback than interactive PB-SMT.
Due to these large differences, in the rest of this work, we focus only on neural systems.

\subsubsection{Translation post-editing with online learning}

Now, we study the application of online learning during the post-editing stage. We built adaptive NMT systems, able to continuously learn from previous errors. Hopefully, this will lead to better systems, with higher translation quality and with a subsequent decrease of the effort required for correcting their outputs.

As we have available validation data, we conducted an exploration of the best SGD optimizer and its hyperparamters for each task. We performed a grid search over the validation set and chose the configurations which obtained the lowest TER values. We compared vanilla SGD, Adagrad, Adadelta and Adam algorithms. We left the learning rate as the only tunable hyperparameter. The rest were fixed to default. We explored learning rates in the values: $b \cdot 10 ^{e}, b\in\{1, 5\}, e \in \{1, -1, -2, -3, -4, -5, -6\}$.

\begin{wraptable}{r}{6cm}
	\caption{\label{table:results-optimizer-scenario1}Best online SGD optimizer for each task.}
	\centering	
	\small
	\begin{tabular}{lllr}
		\toprule
		& & Algorithm & $\rho$ \\
		\midrule
		\multicolumn{1}{ c }{\multirow{2}{*}{XRCE} } &
		En$\rightarrow$De  & Adadelta & $0.1$\\ 
		\multicolumn{1}{ c  }{}                        &
		En$\rightarrow$Fr    & Adadelta & $0.1$ \\
		\midrule
		\multicolumn{1}{ c }{\multirow{2}{*}{EU} } &
		En$\rightarrow$De  & SGD & $10^{-5}$\\ 
		\multicolumn{1}{ c  }{}                        &
		En$\rightarrow$Fr    & Adadelta & $0.1$ \\ 
		\midrule
		\multicolumn{1}{ c }{\multirow{2}{*}{UFAL} } &
		En$\rightarrow$De  & Adadelta & $0.1$\\ 
		\multicolumn{1}{ c  }{}                        &
		En$\rightarrow$Fr & Adadelta & $0.1$\\ 
		\midrule
		\multicolumn{1}{ c }{\multirow{2}{*}{Europarl} } &
		En$\rightarrow$De  & SGD &  $0.005$\\ 
		\multicolumn{1}{ c  }{}                        &				
		En$\rightarrow$Fr  & Adadelta & $0.1$\\ 
		\midrule
		\multicolumn{1}{ c }{\multirow{2}{*}{TED} } &
		En$\rightarrow$De  & Adadelta & $0.1$\\ 
		\multicolumn{1}{ c  }{}                        &
		En$\rightarrow$Fr  & SGD & $0.01$\\ 
		\bottomrule
	\end{tabular}
\end{wraptable}
\cref{table:results-optimizer-scenario1} shows the best configuration obtained for each task. Adadelta outperformed other algorithms, for most tasks. This finding is contrary to the observed by~\citet{Turchi17}, who found that the most effective algorithm was vanilla SGD. We conjecture that these differences are due to the fact that they fixed the learning rate of all dynamic optimizers to $0.01$, which may be unsuitable in some cases.

In our experimentation, Adadelta was the most stable optimizer, obtaining the best results using learning rates of $0.1$ or $0.5$. Vanilla SGD also performed well, but we found it to be more unstable than Adadelta: it was harder to find an adequate learning rate and it performed worse than Adadelta in many cases. On the other hand, we found that Adagrad and Adam were excessively aggressive to be useful for adaptive NMT systems: they required very low learning rates, otherwise, they completely distort the model. And even with low learning rates, they always performed worse than Adadelta or plain SGD.

\cref{table:results-OL} shows the effect of including OL in the post-editing process. In almost every case, the effort required, measured in terms of TER, is improved, yielding significant reductions in all tasks but EU. BLEU is consequently improved, following the TER trend.

\begin{table}[!h]
	\caption{\label{table:results-OL}
		Translation results in terms of TER [\%] and BLEU [\%], of an adaptive NMT system (OL-NMT), compared to static NMT. $\nabla$ and $\Delta$ represent absolute decrements and increments in terms of percentage for the corresponding metric of the online NMT system with respect the static one. Results that are statistically significantly better for each task and metric are boldfaced. }  
	\centering
	\small
	\begin{tabular}{llllrllr}
		\toprule
		& & \multicolumn{3}{c}{TER [\%]} 
		& \multicolumn{3}{c}{ BLEU [\%]}  
		\\ \cmidrule(lr){3-5}\cmidrule(lr){6-8} 
		& &NMT& OL-NMT & $\nabla$ &  NMT& OL-NMT& $\Delta$\\
		\midrule
		\multicolumn{1}{ c }{\multirow{2}{*}{XRCE} } &
		En$\rightarrow$De  & $64.1 \pm 1.1$  & $\mathbf{60.0 \pm 1.1}$  & $4.1$ & $24.4 \pm 1.1$ & $\mathbf{28.9 \pm 1.2}$ & $4.5$ \\ 
		\multicolumn{1}{ c  }{}                        &
		En$\rightarrow$Fr    & $54.0 \pm 1.0$ & $\mathbf{48.7 \pm 1.1}$& $5.3$ & $34.7 \pm 1.2$ & $\mathbf{40.3 \pm 1.3}$& $5.6$\\ 
		\midrule
		\multicolumn{1}{ c }{\multirow{2}{*}{EU} } &
		En$\rightarrow$De    & $54.6 \pm 1.0$  &  $53.8 \pm 1.0$  & $0.8$ & $35.8 \pm 1.1$  & $36.1 \pm 1.1$ & $0.3$ \\ 
		\multicolumn{1}{ c  }{}                        &
		En$\rightarrow$Fr   & $39.3 \pm 0.8$ & $39.3 \pm 0.8$ & $0.0$ & $50.0 \pm 0.9$  & $50.2 \pm 0.9$ &$0.2$\\ 
		\midrule
		\multicolumn{1}{ c }{\multirow{2}{*}{UFAL} } &
		En$\rightarrow$De    & $56.5 \pm 0.6$  & $\mathbf{53.7\pm 0.6}$ &$2.8$ & $24.2 \pm 0.5$  & $\mathbf{26.0\pm 0.6}$  &$1.8$\\ 
		\multicolumn{1}{ c  }{}                        &
		En$\rightarrow$Fr   & $46.4 \pm 0.6$ & $\mathbf{41.7  \pm 0.6}$ &$4.7$ & $37.2 \pm 0.6$  & $\mathbf{41.9 \pm 0.6} $&  $4.7$\\ 
		\midrule
		\multicolumn{1}{ c }{\multirow{2}{*}{Europarl}} &
		
		En$\rightarrow$De   & $ 63.1 \pm 0.4 $ & $\mathbf{60.4 \pm 0.4}$  &$2.7$ & $ 20.0 \pm 0.3 $  & $\mathbf{22.8\pm 0.3}$  &$2.8$\\ 			
		\multicolumn{1}{ c  }{}                        &				
		En$\rightarrow$Fr  & $55.0 \pm 0.3$ & $\mathbf{53.4 \pm 0.3}$  &$1.6$& $27.8 \pm 0.3$  & $\mathbf{29.7  \pm  0.3}$ &$1.9$\\ 
		
		\midrule
		\multicolumn{1}{ c }{\multirow{2}{*}{TED} } &
		En$\rightarrow$De    & $55.5 \pm 0.6$ & $\mathbf{50.8 \pm 0.5}$ & $4.7$&  $24.5 \pm 0.5$ & $\mathbf{25.5  \pm 0.5}$& $1.0$\\ 
		\multicolumn{1}{ c  }{}                        &
		En$\rightarrow$Fr   & $51.5 \pm 0.5$ & $\mathbf{50.5\pm 0.5} $  &$1.0$ & $32.1 \pm 0.5$  & $32.9 \pm 0.5$ &$0.7$\\ 
		\bottomrule
	\end{tabular}
\end{table}

The largest improvements were obtained in the XRCE task, with gains of $4.1$ and $5.3$ TER points. This was due to the high RR, RRR and UNF values of this task (\cref{table:corpora}), being especially adequate for incremental learning. We also observed significant gains in the UFAL, TED and Europarl tasks.

On the other hand, in the EU task, the adaptive system achieved very little differences with respect to the static one. The low values of RR, RRR (almost zero) and UNF explain this little enhancement. Given the lack repetitiveness of the document, an online system is unable to exploit the recently learned knowledge. \label{sec:excusa-ue}

Compared to the existing literature, \citet{Ortiz16}, applied OL techniques to a PB-SMT system, for the XRCE and Europarl tasks. In the first case, they obtained large improvements in terms of BLEU ($11.5$ and $8.4$, for En$\rightarrow$De and En$\rightarrow$Fr, respectively). Nevertheless, their baseline systems performed worse than ours. Including OL, they achieved a similar performance for the XRCE task than our online NMT systems. In the Europarl case, they improved their static system by $1.0$ and $1.4$ BLEU points, for En$\rightarrow$De and En$\rightarrow$Fr. But in this case, their static systems were much worse than ours ($13.1$/$21.2$, for En$\rightarrow$De/En$\rightarrow$Fr). NMT systems perform better than PB-SMT if they have a large amount of training data~\citep{Koehn17}, as in this case. Even though, OL was able to refine a strong NMT system trained with a large corpus.

\subsubsection{Interactive machine translation with online learning}

\begin{table}[!h]
	\caption{\label{table:results-task-char-INMT} KSMR~[\%] effort required for adaptive NMT systems (OL-INMT) compared to static INMT and the state-of-the-art IMT adaptive systems (based on PB-SMT). Results that are statistically significantly better for each task and metric are boldfaced. $^\dagger$ refers to \citet{Ortiz16} and $-$ indicates missing results in the literature.}
	\centering
	\small
	\begin{tabular}{llrrr}
		\toprule
		& & \multicolumn{3}{c}{KSMR [\%]} 
		\\ \cmidrule(lr){3-5}
		& &  INMT& OL-INMT & Best in literature \\
		\midrule
		\multicolumn{1}{ c }{\multirow{2}{*}{XRCE} } &
		En$\rightarrow$De  & $ 32.2 \pm 0.7$ & $\mathbf{27.9 \pm 0.7}$& $37.0 \pm 1.3$$^\dagger$   \\
		\multicolumn{1}{ c  }{}                        &
		En$\rightarrow$Fr & $ 27.5 \pm 0.6$ & $\mathbf{22.5\pm 0.6}$& $30.3 \pm 1.2$$^\dagger$  \\
		\midrule
		\multicolumn{1}{ c }{\multirow{2}{*}{EU} } &
		En$\rightarrow$De  & $19.8 \pm 0.5$ &  $19.5 \pm 0.5$& $-$ \\
		\multicolumn{1}{ c  }{}                        &
		En$\rightarrow$Fr  & $16.3 \pm 0.4$ & $16.2 \pm 0.4$ & $-$  \\
		\midrule
		\multicolumn{1}{ c }{\multirow{2}{*}{UFAL} } &
		En$\rightarrow$De   & $ 22.5 \pm 0.3$ & $\mathbf{21.5\pm 0.3}$& $-$   \\
		\multicolumn{1}{ c  }{}                        &
		En$\rightarrow$Fr& $ 19.7 \pm 0.3$ & $\mathbf{17.7 \pm 0.3}$  & $-$  \\
		\midrule
		\multicolumn{1}{ c }{\multirow{2}{*}{Europarl} } &
		En$\rightarrow$De    & $ 32.9 \pm 0.2 $ & $\mathbf{29.0 \pm 0.2}$& $48.0\pm 0.5$$^\dagger$ \\
		\multicolumn{1}{ c  }{}                        &
		En$\rightarrow$Fr  & $ 29.8 \pm 0.2 $ & $\mathbf{28.0 \pm0.2 }$& $43.2 \pm 0.5$$^\dagger$  \\
		\midrule
		\multicolumn{1}{ c }{\multirow{2}{*}{TED} } &
		En$\rightarrow$De  & $28.0 \pm 0.3$ & $27.5\pm0.3$  & $-$ \\
		\multicolumn{1}{ c  }{}                        &
		En$\rightarrow$Fr& $26.8 \pm 0.3$ & $\mathbf{26.2\pm 0.3}$  & $-$  \\
		\bottomrule
	\end{tabular}
\end{table}

Next, we move towards the deployment of adaptive, interactive NMT systems. We used the same configuration as for post-editing. \cref{table:results-task-char-INMT} shows the effect of adding OL to INMT systems. Moreover, we show other results obtained in the literature.

We found that adaptive INMT systems outperformed static ones. As in post-editing, most of these differences were significant. Again, the XRCE task is the most benefited by OL, but we also obtained especially good results in the Europarl corpora. Besides their RRR and UNF values, it should also be noticed that the Europarl test documents had more samples than others. Therefore, the INMT system was benefited from a longer process of adaptation.

The EU task had the same behavior than in post-editing (\cref{table:results-OL}): because of the lack of repetitiveness, we obtained marginal and non-significant improvements. 

Compared to the literature~\citep{Ortiz16}, we obtained similar gains in terms of KSMR for the XRCE task (around $5$ KSMR points). In the case of Europarl, we obtained higher KSMR decreases: $3.9$/$1.8$ against $1.2$/$1.2$, for the En$\rightarrow$De and En$\rightarrow$Fr language pairs, respectively. Moreover, the large advantage in KSMR that INMT systems had with respect PB-SMT models is maintained in the online version.

\subsection{Scenario \#2: Lack of in-domain data}
\label{sec:scenario2}

In this second scenario, we assume that we have no in-domain training data available. This can be the case of a system trained with data from a general domain, but having to translate documents from a different (and potentially unknown) domain. We take advantage of OL for performing domain adaptation on-the-fly, from the general to the test domain. We expect to obtain better system hypotheses as the post-editing, and inherently the online learning processes go on. The refinement of the system will hopefully entail a decrease of the human effort required for post-editing the upcoming samples. 
	
We took as general system the one trained with the Europarl corpus. In order to work with the same vocabulary, we applied the same BPE segmentation to all in-domain sets. \cref{table:vocabulary_intersection} shows the vocabulary coverage of the NMT system and each one of the in-domain documents and the RRR and UNF metrics for each task. All values were computed according to the BPE version of each test document.

\begin{table}[!h]
		\centering
		\caption{\label{table:vocabulary_intersection}Vocabulary coverage ($C$), restricted repetition rate (RRR) and unseen $n$-gram fraction (UNF) with respect to the out-of-domain corpus (Europarl). We applied the BPE codification learned for the Europarl corpus to each document.}  
		{   %\renewcommand{\arraystretch}{1.1}
			\small
            \begin{tabular}{llccrrr}
                     	\toprule
                     	& & Training
                     	& Development
                     	& \multicolumn{3}{c}{Test}  \\
                     	\cmidrule(lr){3-3}\cmidrule(lr){4-4}  \cmidrule(lr){5-7}
                     	&  & $C$ & $C$ & $C$ &RRR$~[\%]$ &UNF$~[\%]$\\
                     	\midrule
                     	\multirow{4}{*}{XRCE} &
                     	De   & $98.7$ & $99.8$ & $99.5$ & $26.8$ & $12.9$ \\
                     	& En  & $99.7$ & $99.9$ & $99.9$ & $27.7$ & $11.7$\\
                     	\cmidrule(lr){2-7}
                     	& Fr  & $98.2$ & $97.5$ & $97.4$  & $28.6$ & $8.8$\\
                     	& En  & $98.5$ & $97.2$ & $97.2$  & $33.1$ & $8.4$\\
                     	\midrule
                     	\multirow{4}{*}{EU} &
                     	De   & $98.2$ & $99.9$ & $99.9$ & $7.6$ & $19.5$\\
                     	& En  & $95.1$ & $99.7$ & $99.6$ & $8.1$ & $17.5$ \\
                     	\cmidrule(lr){2-7}
                     	& Fr  & $97.5$ & $98.0$ & $98.6$  & $9.9$ & $15.3$ \\
                     	& En  & $96.0$ & $97.6$ & $98.4$  & $8.2$ & $17.2$ \\
                     	\midrule
                     	\multirow{4}{*}{UFAL} &
                     	De   & $92.2$ & $99.8$ & $99.7$ & $7.4$ & $8.4$\\
                     	& En  & $85.9$ & $99.8$ & $99.8$ & $13.4$ & $7.8$\\
                     	\cmidrule(lr){2-7}                    
                     	& Fr  & $91.3$ & $99.7$ & $99.7$ & $12.6$ & $7.1$ \\
                     	& En  & $92.5$ & $99.7$ & $99.8$ & $10.5$ & $7.6$ \\
                     	\midrule
                     	\multirow{4}{*}{TED}&
                     	De   & $98.1$ & $99.8$ & $99.9$ & $4.7$ & $5.4$\\
                     	& En   & $99.2$ & $99.9$ & $99.9$ & $11.0$ & $4.5$ \\
                     	\cmidrule(lr){2-7}                    
                     	& Fr  & $97.9$ & $98.7$ & $99.0$ & $6.4$ & $4.0$\\
                     	& En  & $98.3$ & $98.7$ & $98.8$ & $8.9$ & $3.6$ \\
                     	\bottomrule
            \end{tabular}
		}
	\end{table}

The vocabulary coverage was extremely high for every task (in all cases over $97\%$), showing that BPE can effectively leverage vocabulary differences among domains. 
The RRR and UNF values were increased with respect to the original BPE segmentation (\cref{table:corpora}). This is unsurprising: as we work with different domains, we now have more $n$-grams from our test documents which were unseen in the Europarl training data. Moreover, since rare words tend to be split by the BPE process, it is likely to have more words broken up in the test documents with the BPE from Europarl. Therefore, the repetition rate is increased.
Finally, it is also remarkable that language pairs involving German usually obtained the highest values of UNF. This is because German is more inflective than English or French. Therefore, it is likely to have higher UNF. Since we applied joint BPE, this also affected to the corresponding part in English. 

As we assumed no in-domain data and a potentially unknown domain, we lacked of development sets for this task. Therefore, we used the same algorithm and learning rate for all tasks. We took advantage of the exploration carried out in~\cref{sec:scenario1}. Following~\cref{table:results-optimizer-scenario1}, we applied Adadelta with a learning rate of $0.1$.

\subsubsection{Translation post-editing with online learning}

First, we compare the effort required for post-editing the outputs of the neural system. \cref{table:results-europarl-OL} shows the results of translation quality, in terms of TER and BLEU, for static and adaptive NMT systems. As expected, the translation quality was much lower than in the previous scenario. Differences were especially severe in the XRCE and EU tasks. This is mostly due to the features of each corpora.

The XRCE task relates to printer manuals and contains many short sentences, referring to technical details. Additionally, such manuals usually have formatting templates. Since the system never faced such templates, it made many mistakes. Note that the TER was extremely high in this task. This phenomenon, also observed by \citet{Chinea17} for the same task, is due to the translation of short sentences with a NMT system trained on long sentences from a different domain (Europarl). Therefore, the system generated hypotheses much longer than the references. Therefore, in order to match the reference, TER must delete many words. This problem may be addressed via heuristics in the search method (as pointed out by \citet{Chinea17}), but this is out of the scope of this work.

The EU task is also highly structured. As it records the European Union Bulletin, it contains many sentences from official templates, which have a particular, formal style. It also contains many records with a rigid formatting template, as in the XRCE task. The Europarl corpus mostly records speeches given in the European Parliament. Hence, differences among the EU and the Europarl corpora are notorious.

\begin{table}[!h]
	\caption{\label{table:results-europarl-OL}
		Translation post-editing results, in terms of TER [\%] and BLEU [\%], for adaptive NMT systems (OL-INMT) compared to static NMT. The NMT system was trained exclusively on Europarl data. $\nabla$ and $\Delta$ represent absolute decrements and increments in terms of percentage of the corresponding metric. Results that are statistically significantly better for each task and metric are boldfaced. }  
	\centering
	\small
	\begin{tabular}{llllrllr}
		\toprule
		& & \multicolumn{3}{c}{TER [\%]} 
		& \multicolumn{3}{c}{ BLEU [\%]}  
		\\ \cmidrule(lr){3-5}\cmidrule(lr){6-8} 
		& &NMT& OL-NMT & $\nabla$&  NMT& OL-NMT & $\Delta$\\
		\midrule
		\multicolumn{1}{ c }{\multirow{2}{*}{XRCE} } &
		En$\rightarrow$De  & $ 86.2 \pm 0.9 $ & $\mathbf{68.3 \pm 1.1}$ & $17.9$ & $ 6.6 \pm 0.5$ & $ \mathbf{20.4 \pm 0.9}$ & $13.8$ \\ 
		\multicolumn{1}{ c  }{}                        &
		En$\rightarrow$Fr      & $76.3 \pm 1.1$ & $\mathbf{58.9 \pm 1.0}$ & $17.4$ & $ 12.8\pm 0.6$ & $\mathbf{29.1 \pm 1.0}$ & $16.3$\\ 
		\midrule
		\multicolumn{1}{ c }{\multirow{2}{*}{EU} } &
		En$\rightarrow$De   & $ 73.5 \pm 0.9$ & $\mathbf{69.4 \pm 0.9}$ & $4.1$ & $ 18.1 \pm 0.6$ & $\mathbf{20.8 \pm 0.6}$ & $2.7 $ \\ 
		\multicolumn{1}{ c  }{}                        &
		En$\rightarrow$Fr     & $ 59.6 \pm 0.7$ & $\mathbf{56.5 \pm 0.9}$ & $3.1$ & $27.7 \pm 0.6$ & $\mathbf{33.2 \pm 0.6}$ & $5.5$\\ 
		\midrule
		\multicolumn{1}{ c }{\multirow{2}{*}{UFAL} } &
		En$\rightarrow$De  & $ 66.9 \pm 0.6$ & $\mathbf{62.8 \pm 0.6}$ & $4.1$ & $15.7 \pm 0.4$ & $\mathbf{18.8 \pm 0.4}$ & $3.1$ \\ 
		\multicolumn{1}{ c  }{}                        &
		En$\rightarrow$Fr      & $ 52.9 \pm 0.6$ & $\mathbf{49.9 \pm 0.6}$ & $ 3.0 $ & $ 29.5 \pm 0.5$ & $\mathbf{33.4 \pm 0.5}$ & $ 3.9 $\\ 
		\midrule
		\multicolumn{1}{ c }{\multirow{2}{*}{TED} } &
		En$\rightarrow$De  & $ 61.5 \pm 0.5$ & $\mathbf{56.8 \pm 0.5}$ & $ 4.7 $ & $20.4\pm 0.4$ & $\mathbf{23.4\pm 0.5}$ & $3.0 $ \\ 
		\multicolumn{1}{ c  }{}                        &
		En$\rightarrow$Fr & $ 57.3 \pm 0.5$ & $\mathbf{52.7 \pm 0.5}$ & $4.6$ & $27.3 \pm 0.5$ & $ \mathbf{31.1 \pm 0.5}$ & $3.8$\\  
		\bottomrule
	\end{tabular}
\end{table}

On the other hand, the UFAL and TED corpora are closer to Europarl. Although the domains are different (medical and talks), the style, constructions and template of these documents are similar to Europarl. Therefore, the differences in terms of translation quality are smaller. \cref{fig:XeroxEUexamples} shows examples of common sentences from all four tasks.

\begin{figure}[h!]
	\caption{Examples of the XCRE, EU, UFAL and TED tasks, in English. All sentences belong to the corresponding test set. Sentences from the XRCE and EU corpora are highly structured, while the other tasks have a more natural style.\label{fig:XeroxEUexamples}}
	\centering
	\scriptsize
	\begin{tabular}{l l}
		\toprule
		\multirow{2}{*}{XRCE} &\textit{* press " select " to save the setting .} \\
		&  \textit{* press " output " to select " on " .}  \\
		\cmidrule(lr){1-2}  
		\multirow{2}{*}{EU} &  \textit{31996 Y 0801 ( 04 ) Council Resolution of 15 July 1996 on the transparency of vocational training certificates .} \\
		&  \textit{Reference : Conclusions of the Vienna European Council : Bull. 12-1998 , points I. 19 et seq.} \\
		\cmidrule(lr){1-2}  		
		\multirow{2}{*}{UFAL} &  \textit{It 's a long , hollow tube at the end of your digestive tract where your body makes and stores stool .} \\
		&  \textit{We are also studying how their work affects the quality of their lives .} \\
		\cmidrule(lr){1-2}  		
		\multirow{2}{*}{TED} &  \textit{Everybody talks about happiness these days .
		} \\
		&  \textit{I 'm going to talk today about energy and climate .} \\
\midrule\midrule

		\multirow{2}{*}{Europarl} &  \textit{A Republican strategy to counter the re-election of Obama 
		} \\
		&  \textit{Republican leaders justified their policy by the need to combat electoral fraud .} \\
		\bottomrule
	\end{tabular}
\end{figure}

As expected, the development of adaptive NMT systems greatly improved the quality of the systems. The improvements brought by continuous learning to the XRCE task were large ($17.9$/$17.4$ points in terms of TER and $13.8$/$16.3$ in terms of BLEU). These large improvements were due to the aforementioned structure features of this text. Since the text was extracted from printer manuals, the restricted repetition rate was extremely high: more than a $25\%$ in al cases (see \cref{table:vocabulary_intersection}). 

OL is more effective in texts with high RRR, since upcoming events have been already seen. This effectiveness is boosted by our experimental conditions: we were translating with a general NMT system. Therefore, the systems was prone to make the same errors over and over. As we introduced continuous learning in the system, it rapidly adapted to the XRCE features. Since the XRCE test set is quite repetitive, the OL-NMT avoided to make the same error again, which had a great impact in effort reduction and translation quality. 

OL was also effective for the rest of tasks, obtaining consistent improvements, ranging from $3.0$ to $4.7$ TER points. It is worth noting that for the TED (En$\rightarrow$De) task, a system exclusively trained on out-of-domain data and fine-tuned via OL, was able to outperform a PB-SMT system trained on in-domain data (\cref{table:results-mt-quality}). The rest of OL systems also behaved good, achieving performances close to the systems trained on in-domain data. These results demonstrate that online learning is a good choice when developing translation systems with scarce data resources.

This experiment is comparable to the \textit{a posteriori} adaptation strategy developed by \citet{Turchi17}. They also adapted a general model to a given domain by means of incremental learning on post-edited samples, obtaining significant BLEU improvements.

\subsubsection{Interactive machine translation with online learning}

Next, we study the effectiveness of the general NMT system in the interactive framework and the effect of OL-based adaptation in terms of effort reduction. 
\cref{table:results-europarl-char-INMT} shows the INMT results of an adaptive system and a static one. 

\begin{table}[!ht]
	\caption{\label{table:results-europarl-char-INMT}Human effort required for INMT systems with online learning (OL-INMT), compared to static INMT, in terms of KSMR [\%]. NMT systems were exclusively trained on Europarl data. $\nabla$ represents absolute decrements in percentage. Results that are statistically significantly better for each task and metric are boldfaced.}  
	\centering
	\small
	\begin{tabular}{llllr}
		\toprule
		& & \multicolumn{3}{c}{KSMR [\%]} 
		\\ \cmidrule(lr){3-5}
		& &INMT& OL-INMT & $\nabla$ \\
		\midrule
		\multicolumn{1}{ c }{\multirow{2}{*}{XRCE} } &
		En$\rightarrow$De  & $ 52.2 \pm 0.6$ & $\mathbf{35.1 \pm 0.7 } $ & $17.1$ \\
		\multicolumn{1}{ c  }{}                        &
		En$\rightarrow$Fr    & $ 56.6 \pm 0.7 $ & $\mathbf{32.9 \pm 0.7}$ & $23.7$ \\
		\midrule
		\multicolumn{1}{ c }{\multirow{2}{*}{EU} } &
		En$\rightarrow$De & $28.0 \pm 0.4 $ &  $\mathbf{26.4 \pm 0.5}$  & $1.6$\\
		\multicolumn{1}{ c  }{}                        &
		En$\rightarrow$Fr   & $27.0 \pm 0.5$ & $\mathbf{23.2\pm 0.4}$ & $3.8$ \\
		\midrule
		\multicolumn{1}{ c }{\multirow{2}{*}{UFAL} } &
		En$\rightarrow$De    & $33.3\pm 0.4$ & $\mathbf{29.7 \pm 0.3}$ & $3.6$ \\
		\multicolumn{1}{ c  }{}                        &
		En$\rightarrow$Fr & $ 29.6 \pm 0.3 $ & $\mathbf{25.6 \pm 0.3}$ & $4.0$ \\
		\midrule
		\multicolumn{1}{ c }{\multirow{2}{*}{TED} } &
		En$\rightarrow$De   & $30.2 \pm 0.3$ & $\mathbf{28.0 \pm 0.3}$ & $2.2$ \\
		\multicolumn{1}{ c  }{}                        &
		En$\rightarrow$Fr& $ 28.3 \pm 0.3$ & $\mathbf{26.0\pm 0.3}$ & $2.3$ \\
		\bottomrule
	\end{tabular}
\end{table}

The performance drop of the general INMT system depended on the task: in tasks with domains close to the general corpus, performances of general INMT systems were close to an in-domain system (e.g. TED). But, if the domain of the test data is far from  the general corpus (XRCE), the human effort required dramatically rose.

The introduction of OL into the interactive systems, had a similar effect to that observed in \cref{table:results-europarl-OL}: we obtained significant KSMR reductions for all tasks. The greatest improvements were again obtained in the XCRE task, due to the aforementioned reasons (highest RRR and shortest sentences). In the case of the TED task, online learning overcame the gap between training a specific system or using the general one. Interestingly, an adaptive INMT system, trained on out-of-domain data, performed better than PB-SMT state-of-the-art systems (\cref{table:results-task1-literature}).

\subsection{Scenario \#3: Fine-tuning a general neural machine translation system}

In our last experimental setup, we hybridized scenarios \#1 and \#2: we have available in-domain and out-of-domain data. Thus, we started from a general NMT system, trained on an out-of-domain corpus, and fine-tuned it with the in-domain training data. Finally, we followed the OL refinement procedure, as in previous scenarios. We study if OL can bring enhancements to an already fine-tuned system, and if so, to what extent.

Again, we used the Europarl corpus as out-of-domain. We followed the same segmentation strategy than in \cref{sec:scenario2}, applying it also to the training set. \cref{table:vocabulary_intersection} shows the vocabulary coverage of the training sets, generally high. Only in the UFAL corpus the coverage was slightly lower (from $85.9$\% to $92.5$\%). 

Once we had our system trained on Europarl, we continued the training on each in-domain training set. For this retraining, we kept the hyperparameters used for training the original NMT system: Adam with $\rho = 0.0002$. Following \citet{Wu16}, we also tested vanilla SGD with learning rate annealing, but we obtained poorer results. We early-stopped the training following the same criterion as in the general case (\cref{sec:training_systems}), but setting the patience to $10$ evaluations. 

\begin{wraptable}{r}{6.5cm}
	\caption{\label{table:results-optimizer-scenario3}Best online SGD optimizer for each task for fine-tuned NMT systems.}
	\centering	
	\small
	\begin{tabular}{lllr}
		\toprule
		& & Algorithm & $\rho$ \\
		\midrule
		\multicolumn{1}{ c }{\multirow{2}{*}{XRCE} } &
		En$\rightarrow$De  & Adadelta & $0.5$\\ 
		\multicolumn{1}{ c  }{}                        &
		En$\rightarrow$Fr    & Adadelta & $0.5$ \\
		\midrule
		\multicolumn{1}{ c }{\multirow{2}{*}{EU} } &
		En$\rightarrow$De  & SGD & $10^{-4}$\\ 
		\multicolumn{1}{ c  }{}                        &
		En$\rightarrow$Fr    & Adadelta & $ 0.1$ \\ 
		\midrule
		\multicolumn{1}{ c }{\multirow{2}{*}{UFAL} } &
		En$\rightarrow$De  & Adadelta & $0.5$\\ 
		\multicolumn{1}{ c  }{}                        &
		En$\rightarrow$Fr & Adadelta & $0.5$\\ 
		\midrule
		\multicolumn{1}{ c }{\multirow{2}{*}{TED} } &
		En$\rightarrow$De  &Adadelta & $0.1$ \\ 
		\multicolumn{1}{ c  }{}                        &
		En$\rightarrow$Fr  & SGD & $0.01$\\ 
		\bottomrule
	\end{tabular}
\end{wraptable} 

We repeated the same grid search exploration for obtaining the best hyperparameters for each OL optimizer, as in \cref{sec:scenario1}. Interestingly, the top-performing algorithms were the same in both scenarios. Adadelta or SGD obtained the best performance. But in this case, they required higher learning rates. \cref{table:results-optimizer-scenario3} shows the best configuration for each task.

\subsubsection{Translation post-editing with online learning}

As in previous sections, we first compare static and adaptive systems, in terms of translation quality and effort required in the post-editing process. 
\cref{table:results-task-europarl-ONMT} shows the results of the systems initialized with Europarl and fine-tuned with their corresponding in-domain training data.

\begin{table}[!ht]
	\caption{\label{table:results-task-europarl-ONMT} Translation post-editing results, in terms of TER [\%] and BLEU [\%], for adaptive NMT systems (OL-INMT) compared to static NMT. All NMT systems have been pretrained on Europarl data and fine-tuned with the training data from each task. $\nabla$ represents absolute decrements in terms of percentage of the corresponding metric. Results that are statistically significantly better for each task and metric are boldfaced. }  
	\centering
	\small
	\begin{tabular}{llllrllr}
		\toprule
		& & \multicolumn{3}{c}{TER [\%]} 
		& \multicolumn{3}{c}{ BLEU [\%]}  
		\\ \cmidrule(lr){3-5}\cmidrule(lr){6-8} 
		& &NMT& OL-NMT & $\nabla$ &  NMT& OL-NMT & $\Delta$\\
		\midrule
		\multicolumn{1}{ c }{\multirow{2}{*}{XRCE} } &
		En$\rightarrow$De  & $60.0 \pm 1.0$ & $\mathbf{53.1 \pm1.1}$ & $6.9$ & $27.3\pm 1.2$ & $\mathbf{33.6 \pm 1.2}$ & $6.3$ \\ 
		\multicolumn{1}{ c  }{}                        &
		En$\rightarrow$Fr      & $ 50.1 \pm 1.0$ & $\mathbf{43.2 \pm 1.1}$ & $6.9$ & $38.8 \pm 1.2 $ & $\mathbf{46.6 \pm 1.2}$ & $7.8$ \\ 
		\midrule
		\multicolumn{1}{ c }{\multirow{2}{*}{EU} } &
		En$\rightarrow$De & $51.8 \pm 1.0$ & $51.7 \pm 1.0$ & $0.1$ & $37.1 \pm 1.0$ & $36.8 \pm 1.1$ & $-0.3$ \\ 
		\multicolumn{1}{ c  }{}                        &
		En$\rightarrow$Fr   & $37.9 \pm 0.8$ & $38.2 \pm 0.8$ 	& $0.0$ & $50.6 \pm 0.7$ & $51.8 \pm 0.9$  & $0.2$ \\ 
		\midrule
		\multicolumn{1}{ c }{\multirow{2}{*}{UFAL} } &
		En$\rightarrow$De    & $56.5 \pm 0.6$ & $\mathbf{53.3\pm0.6}$ & $3.2$ & $23.4 \pm 0.6$ & $\mathbf{26.0 \pm 0.6}$ & $2.6$ \\ 
		\multicolumn{1}{ c  }{}                        &
		En$\rightarrow$Fr & $47.8 \pm 0.6$ & $\mathbf{40.9 \pm0.5}$ & $6.9$ & $36.6\pm 0.6$ & $ \mathbf{42.8 \pm0.6}$ & $6.2$ \\ 
		\midrule
		\multicolumn{1}{ c }{\multirow{2}{*}{TED} } &
		En$\rightarrow$De & $52.2 \pm 0.5$ & $\mathbf{51.0\pm 0.5}$ & $1.2$ & $27.5\pm0.6$ & $ \mathbf{29.8 \pm 0.6}$ & $2.3$ \\ 
		\multicolumn{1}{ c  }{}                        &
		 En$\rightarrow$Fr& $ 47.5 \pm 0.5$ & $46.8 \pm 0.5$ & $0.7$ & $36.4\pm 0.5$ & $37.0 \pm0.5$ & $0.6$ \\ 
		\bottomrule
	\end{tabular}
\end{table}

Compared to systems exclusively trained on in-domain data (\cref{table:results-mt-quality}), we found that those initialized from Europarl performed generally better if we have scarce in-domain data. In the XRCE and TED tasks, both TER and BLEU were significantly improved when fine-tuning the general system. We found improvements of approximately $4$ TER points in each task. On the other hand, if we had available a large amount of in-domain data, the fine-tuning effectiveness is diluted. This is the case of the UFAL task: as this is a large in-domain corpus, to pre-train with Europarl had minor effects on the final NMT performance.

The addition of OL to the NMT systems generally improved the performance with respect to static systems. In the XRCE and UFAL tasks, the improvements were large: From $3.2$ to $6.9$ TER points. According to their RRR and UNF metrics (\cref{table:vocabulary_intersection}), this was expected, because those are the corpora with higher RRR. The TED task was also benefited from OL, but to a lower extent. Even though, we observed significant improvements in the En$\rightarrow$De.

On the other hand, in the EU task we obtained almost no differences between static and adaptive systems. This lack of improvement was a common pattern in the UE along all the experimentation conducted in this work. As stated in \cref{sec:excusa-ue}, this is because the RRR and UNF values of this corpus. OL was unable to be exploited to the full in this corpus.

\subsubsection{Interactive machine translation with online learning}

\cref{table:results-task-char-europarl-task3-INMT} shows the effort required in a IMT scenario. Compared to the systems exclusively trained on the in-domain data (\cref{table:results-task-char-INMT}), we observed the same phenomenon as in the previous section: the usage of out-of-domain data was especially effective in tasks with scarce in-domain data. Fine-tuned systems performed clearly better in all cases but UFAL (En$\rightarrow$De). 

\begin{table}[!ht]
	\caption{\label{table:results-task-char-europarl-task3-INMT} Effort required for adaptive NMT systems (OL-INMT) compared to static NMT, in terms of KSMR~[\%]. All NMT systems have been pretrained on Europarl data and fine-tuned with the training data from each task. $\nabla$ represents absolute decrements in terms of percentage. Results that are statistically significantly better for each task and metric are boldfaced. }  
	\centering
	\small
	\begin{tabular}{llllr}
		\toprule
		& & \multicolumn{3}{c}{KSMR [\%]} 
		\\ \cmidrule(lr){3-5}
		& &INMT& OL-INMT & $\nabla$ \\
		\midrule
		\multicolumn{1}{ c }{\multirow{2}{*}{XRCE} } &
		En$\rightarrow$De  & $23.3 \pm 0.6$ & $\mathbf{20.3 \pm 0.6} $ & $3.0$ \\
		\multicolumn{1}{ c  }{}                        &
		En$\rightarrow$Fr & $ 22.4 \pm 0.5$ & $\mathbf{17.8 \pm 0.5}$ & $4.6$ \\
		\midrule
		\multicolumn{1}{ c }{\multirow{2}{*}{EU} } &
		En$\rightarrow$De & $16.4\pm 0.4$ &  $16.3 \pm 0.4$  & $0.1$ \\
		\multicolumn{1}{ c  }{}                        &
		En$\rightarrow$Fr   & $14.0 \pm 0.4$ & $13.6 \pm 0.4 $ & $0.3$ \\
		\midrule
		\multicolumn{1}{ c }{\multirow{2}{*}{UFAL} } &
		En$\rightarrow$De    & $22.9 \pm 0.3$ & $\mathbf{21.7 \pm 0.3}$ & $1.2$ \\
		\multicolumn{1}{ c  }{}                        &
		En$\rightarrow$Fr & $ 18.8\pm 0.3$ & $\mathbf{17.2 \pm 0.3}$ & $1.6$ \\
		\midrule
		\multicolumn{1}{ c }{\multirow{2}{*}{TED} } &
		En$\rightarrow$De   & $23.8 \pm 0.3$ & $\mathbf{23.0\pm0.3}$ & $0.8$ \\
		\multicolumn{1}{ c  }{}                        &
		En$\rightarrow$Fr& $21.8 \pm 0.3$ & $\mathbf{21.2 \pm 0.3}$ & $0.6$ \\
		\bottomrule
	\end{tabular}
\end{table}

The largest improvements were obtained in the XRCE and TED tasks, with less training data. In these cases, the enhancements ranged from $4.2$ to $8.9$ KSMR points. The UE task is also improved if we initialize our models with Europarl, obtaining diminishes of $2.3$ and $3.4$ KSMR points. Finally, fine-tuning had a minor effect on the UFAL task, as the in-domain corpus is large enough to build a good INMT system. Nevertheless, pre-training was harmless and we obtained improvements in the En$\rightarrow$Fr language pair. On the other hand, in the En$\rightarrow$De language pair, we obtained a minor degradation of the model ($+0.4\%$ KSMR points).

The addition of OL to these systems had similar effects than in the previous scenarios. Adaptive systems performed significantly better than static systems in all tasks but EU, due to the aforementioned reasons: the RR, RRR and UNF values.

\subsection{Further analyses}
\label{Sec:QualitativeAnalysis}
Finally, we analyze additional aspects of the proposed systems: the effectiveness of the cLSTM units, response times and computational overhead of the adaptation process via online learning and the effectiveness of the proposed vocabulary-masking strategy. Finally, in order to obtain additional insights of the adaptation via OL in NMT, we show and analyze some qualitative examples.

\subsubsection{Evaluating the cLSTM units}

\begin{table}[!th]
	\caption{\label{table:results-regular-LSTM}
		Results of translation quality for all tasks in terms of TER [\%] and BLEU [\%] comparing regular LSTM units and cLSTM units. cLSTM figures are the same than in \cref{table:results-mt-quality}. Results that are statistically significantly better for each task and metric are boldfaced.}  
	\centering
	\small
		\begin{tabular}{llllll}
		\toprule
		& & \multicolumn{2}{c}{TER [\%]} 
		& \multicolumn{2}{c}{ BLEU [\%]}  
		\\ \cmidrule(lr){3-4}\cmidrule(lr){5-6} 
		& &LSTM & cLSTM & LSTM & cLSTM\\
		\midrule
		\multicolumn{1}{ c }{\multirow{2}{*}{XRCE} } &
		En$\rightarrow$De  & $67.1 \pm 1.0$ & $\mathbf{64.1 \pm 1.1}$ & $21.5 \pm 1.1$ & $\mathbf{24.4 \pm 1.1}$\\ 
		\multicolumn{1}{ c  }{}                        &
		En$\rightarrow$Fr   & $57.1 \pm 1.1$ & $\mathbf{54.0 \pm 1.1}$ &  $32.9 \pm 1.2$  & $\mathbf{34.7 \pm 1.2}$ \\ 
		\midrule
		\multicolumn{1}{ c }{\multirow{2}{*}{EU} } &
		En$\rightarrow$De   & $57.5 \pm 1.0$ & $\mathbf{54.6 \pm 1.0}$  & $32.4  \pm 1.1$ & $\mathbf{35.8 \pm 1.1}$  \\ 
		\multicolumn{1}{ c  }{}                        &
		En$\rightarrow$Fr   &  $41.0 \pm 0.8$ & 	$\mathbf{39.3 \pm 0.8}$ & $47.4 \pm 0.9$ & $\mathbf{50.0 \pm 0.9}$  \\ 
		\midrule
		\multicolumn{1}{ c }{\multirow{2}{*}{UFAL} } &
		En$\rightarrow$De  & $58.3 \pm 0.6$  & $\mathbf{56.5 \pm 0.6}$ &  $21.5 \pm 0.5$  & $\mathbf{24.2 \pm 0.5}$ \\ 
		\multicolumn{1}{ c  }{}                        &
		En$\rightarrow$Fr  & $49.7 \pm 0.6$ &$\mathbf{46.4 \pm 0.6}$& $35.2 \pm 0.6$ &  $\mathbf{37.2 \pm 0.6}$  \\ 	
		\midrule
		\multicolumn{1}{ c }{\multirow{2}{*}{Europarl} } &
		
		En$\rightarrow$De  & $67.0 \pm 0.3$ & $\mathbf{63.1 \pm 0.4}$ & $18.1 \pm 0.3$ & $\mathbf{20.0 \pm 0.3}$ \\ 					
		\multicolumn{1}{ c  }{}                        &				
		En$\rightarrow$Fr  & $62.2 \pm 0.3$ & $\mathbf{55.0 \pm 0.3}$ & $25.3 \pm 0.3 $ & $\mathbf{27.8 \pm 0.3}$  \\ 
		
		\midrule
		\multicolumn{1}{ c }{\multirow{2}{*}{TED} } &
		En$\rightarrow$De  & $59.2 \pm 0.5$ & $\mathbf{55.5 \pm 0.6}$ & $22.6 \pm 0.5$ & $\mathbf{24.5 \pm 0.5}$  \\ 
		\multicolumn{1}{ c  }{}                        &
		En$\rightarrow$Fr  & $54.9 \pm 0.5$ & $\mathbf{51.5 \pm 0.5}$ & $29.3 \pm 0.5$ & $\mathbf{32.1 \pm 0.5}$ \\ 
		\bottomrule
	\end{tabular}
\end{table}

We evaluate the performance of the cLSTM unit proposed in \cref{sec:NMT}. To that end, we compare the results obtained using regular LSTM units in the decoder, against using cLSTM ones. We trained the same systems than in \cref{sec:scenario1}, but using standard LSTM units \citep{Gers00} with attention in the decoder. The rest of hyperparameters of the NMT models remained the same.

The results of this experimentation are shown \cref{table:results-regular-LSTM}, in terms of translation quality (TER and BLEU). In all cases, the systems featuring cLSTM units performed better than those with classical LSTM. The differences were consistent and constant across all tasks: cLSTM units increased the BLEU from around $2$ to $3$ points. TER was also improved in all cases, in this case from around $2$ to almost $5$ points.

In addition and despite having more parameters than regular LSTMs, we did not find a significant computational overhead when using cLSTM units, neither in the training and decoding phases. Hence, we conclude that the inclusion of the conditional mechanism to LSTM units was highly positive for the NMT model.

\subsubsection{Temporal costs of interactive, adaptive NMT}

\begin{wraptable}{r}{5.5cm}
	\caption{\label{table:times} Average interaction response time (RT) and learning time (LT) for all tasks.}  
	\centering
	\small
	\begin{tabular}{ll ll}
		\toprule
		& & RT (s) & LT (s)\\ 
		\midrule
		\multicolumn{1}{ c }{\multirow{2}{*}{XRCE} } &
		En$\rightarrow$De  & $0.14$ & $0.09$\\ 
		\multicolumn{1}{ c  }{}                        &
		En$\rightarrow$Fr    & $0.12$ & $0.08$ \\ 
		\midrule
		\multicolumn{1}{ c }{\multirow{2}{*}{EU} } &
		En$\rightarrow$De & $0.28$ & $0.13$ \\ 
		\multicolumn{1}{ c  }{}                        &
		En$\rightarrow$Fr &  $0.31$ & $0.13$ \\ 
		\midrule
		\multicolumn{1}{ c }{\multirow{2}{*}{UFAL} } &
		En$\rightarrow$De &$0.26$ & $0.15$\\ 
		\multicolumn{1}{ c  }{}                        &
		En$\rightarrow$Fr & $0.26$& $0.15$\\ 
		\midrule		
		\multicolumn{1}{ c }{\multirow{2}{*}{Europarl} } &
		En$\rightarrow$De & $ 0.16$ & $0.14$\\ 
		\multicolumn{1}{ c  }{}                        &
		En$\rightarrow$Fr & $0.13$ & $0.14$\\ 
		\midrule
		\multicolumn{1}{ c }{\multirow{2}{*}{TED} } &
		En$\rightarrow$De & $0.23$ & $0.12$\\ 
		\multicolumn{1}{ c  }{}                        &
		En$\rightarrow$Fr & $0.19$ & $0.12$\\ 
		\bottomrule
	\end{tabular}
\end{wraptable}

This work is framed in an interactive protocol, therefore, response times of the system must be adequate for giving the user the feeling of real-time interaction. According to \citet{Nielsen93}, a response time below 0.1 seconds gives the user a feeling of instantaneous reaction. If the response time is between 0.1 and 1 seconds, the user will notice a delay, but his/her flow of though would stay uninterrupted.
Moreover, since we apply OL after interactively translating each sample, the retraining time should also be considered. 

\cref{table:times} shows the response and learning times for each task1\footnote{Experiments executed on a single GeForce GTX 1080 GPU.}. These values refer to the first scenario (\cref{sec:scenario1}). In the other scenarios we used as NMT system the one trained on Europarl. Therefore, this is the reference for those cases.

Response times were around $0.1$ and $0.3$ seconds. These values are close to the $0.1$ seconds specified by \citet{Nielsen93} for having a real-time user feeling. Therefore, users would notice a slight delay on system reactions, but their focus during the interactive translation process will hopefully be unaffected. Nevertheless, we should confirm this by means of an experimentation with real users.

On the other hand, the learning times were kept constant, regardless the task. Learning times were around 0.1 seconds, therefore, differences between adaptive and static systems were almost unnoticeable in terms of usability.

\subsubsection{Impact of online learning}

\begin{figure*}[!h]
	\centering	
	\begin{subfigure}[t]{0.46\textwidth}
		\centering
		\begin{tikzpicture}
	\begin{axis}[
	legend columns=2, 
	legend style={
		legend entries = {Static, Adaptive},
		legend pos = north east,
		legend style={/tikz/column 2/.style={ column sep=5pt,},font=\tiny},
	},
	ylabel={TER [\%]}, 
	xlabel={\# Sentences}, 
	no markers,
	height=0.2\textheight,	
	width=\textwidth,
	enlarge x limits=false,
	cycle list name=color list]	
	\addplot table[x=Sentences,y=Offline, col sep=space] {task1_ter.dat};
	\addplot [mark=*, blue, dashed] table[x=Sentences,y=Online,col sep=space] {task1_ter.dat};
	\end{axis}
\end{tikzpicture}
		\vspace{-1\baselineskip}
		\caption{\label{fig:cum_ter1} Cumulative TER for UFAL En$\rightarrow$De. Systems trained on in-domain data.}
	\end{subfigure}%
	\qquad 	
	\begin{subfigure}[t]{0.48\textwidth}
		\centering
		\begin{tikzpicture}
	\begin{axis}[
	legend columns=2, 
	legend style={
		legend entries = {Static, Adaptive},
		legend pos = north east,
		legend style={/tikz/column 2/.style={ column sep=5pt,},font=\tiny},
	},
	ylabel={KSMR [\%]}, 
	xlabel={\# Sentences}, 
	no markers,
    height=0.2\textheight,	
	width=\textwidth,
	enlarge x limits=false,
	%cycle list name=color list,
	cycle list name=color list]	
		\addplot table[x=Sentences,y=Offline, col sep=space] 	{task1_ksmr.dat};
		\addplot [mark=*, blue, dashed] table[x=Sentences,y=Online,col sep=space] {task1_ksmr.dat};
		\end{axis}
\end{tikzpicture}
		\vspace{-1\baselineskip}
		\caption{\label{fig:cum_ksmr1}  Cumulative KSMR for UFAL En$\rightarrow$De. Systems trained on in-domain data.}
	\end{subfigure}%
	\\\vspace{0.5\baselineskip}		
	\begin{subfigure}[t]{0.46\textwidth}
		\centering
		\begin{tikzpicture}
	\begin{axis}[
		legend columns=2, 
		legend style={
			legend entries = {Static, Adaptive},
			legend pos = north east,
			legend style={/tikz/column 2/.style={ column sep=5pt,},font=\tiny},
		},
		ylabel={TER [\%]}, 
		xlabel={\# Sentences}, 
		no markers,
		height=0.2\textheight,	
		width=\textwidth,
		enlarge x limits=false,
		cycle list name=color list]	
		\addplot table[x=Sentences,y=Offline, col sep=space] 	{task2_ter.dat};
		\addplot [mark=*, blue, dashed] table[x=Sentences,y=Online,col sep=space] {task2_ter.dat};
	\end{axis}
\end{tikzpicture}
		\vspace{-1\baselineskip}		
		\caption{\label{fig:cum_ter2} Cumulative TER for UFAL En$\rightarrow$De. Systems trained on Europarl.}
	\end{subfigure}%
	\qquad 	
	\begin{subfigure}[t]{0.48\textwidth}
		\centering
		\begin{tikzpicture}
	\begin{axis}[
	legend columns=2, 
	legend style={
		legend entries = {Static, Adaptive},
		legend pos = north east,
		legend style={/tikz/column 2/.style={ column sep=5pt,},font=\tiny},
	},
	ylabel={KSMR [\%]}, 
	xlabel={\# Sentences}, 
	no markers,
	height=0.2\textheight,	
	width=\textwidth,
	enlarge x limits=false,
	%cycle list name=color list,
	cycle list name=color list]	
	\addplot table[x=Sentences,y=Offline, col sep=space] 	{task2_ksmr.dat};
	\addplot [mark=*, blue, dashed] table[x=Sentences,y=Online,col sep=space] {task2_ksmr.dat};
	\end{axis}
\end{tikzpicture}
		\vspace{-1\baselineskip}		
		\caption{\label{fig:cum_ksmr2} Cumulative KSMR for UFAL En$\rightarrow$De. Systems trained on Europarl.}
	\end{subfigure}
	
	\caption{Cumulative TER and KSMR of static (solid lines) and adaptive (dashed lines) NMT systems for the UFAL En$\rightarrow$De task. Plots \cref{fig:cum_ter1} and \cref{fig:cum_ksmr1} refer to systems trained on in-domain data, while results of \cref{fig:cum_ter2} and \cref{fig:cum_ksmr2} were obtained with a system trained only on out-of-domain data. \label{fig:cum_ol}}
\end{figure*}
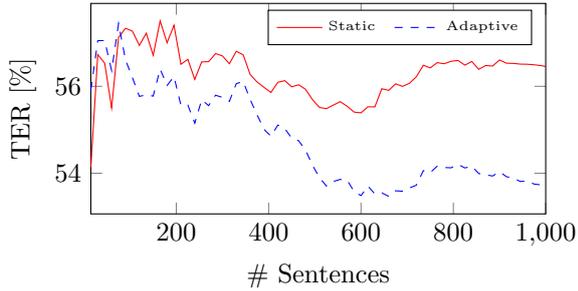
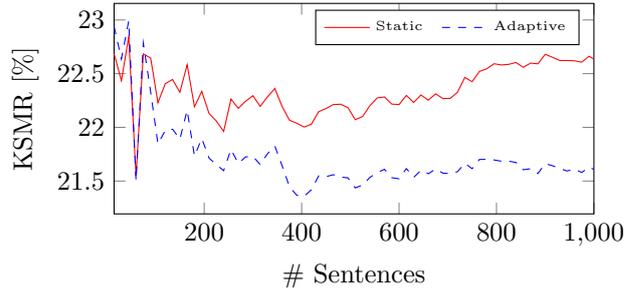
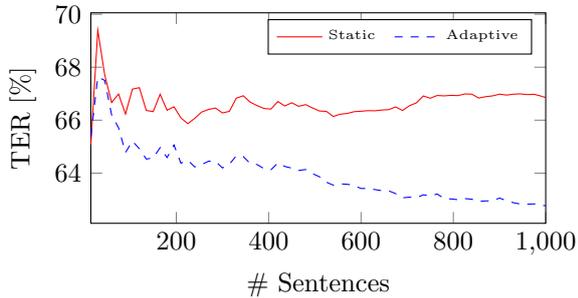
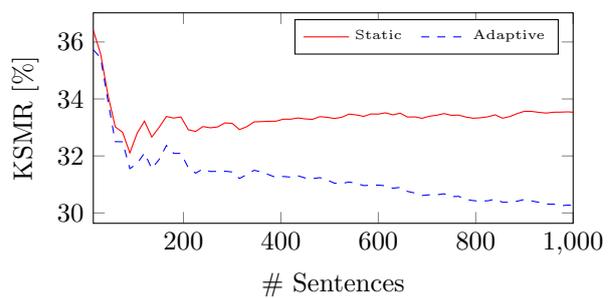

We deepen in the effects of continuous learning in NMT, comparing adaptive versus non-adaptive systems in translation post-editing and IMT. Since we want to reduce the effort required by the user, we are interested in TER and KSMR. We measured cumulative TER and KSMR, as the post-editing and IMT processes advanced. We report results (\cref{fig:cum_ol}) from the UFAL En$\rightarrow$De task, for the systems trained on in-domain and out-of domain data (\cref{sec:scenario1} and \cref{sec:scenario2}, respectively).
 
As shown in \cref{fig:cum_ter1} and \cref{fig:cum_ksmr1}, the adaptive systems were able to rapidly take advantage of the post-edited samples. With approximately 100 samples, TER and KSMR were considerably lowered; with 600 sentences, the differences were large. From here, the systems get on a performance plateau. Nevertheless, if we attend to the static system, we observe that from the sentence 600, the task becomes more difficult, and the TER and KSMR were increased. OL prevented some of this rise, stabilizing the performance of the systems.

OL applied to systems exclusively trained on out-of-domain data (\cref{fig:cum_ter2} and \cref{fig:cum_ksmr2}), improved the performance of the systems. Both TER and KSMR followed a continuous drop. Although expected, this behavior confirms that the systems could be enhanced to larger extents by means of continuous learning, provided that we had more data.

\subsubsection{Vocabulary-masking strategy}

We introduced (\cref{sec:INMT-mask}) a simple yet effective way for performing character-level interactions on a NMT system that works at word (or subword) level. A system with character-level interactions will potentially require less keystrokes than another based on word-level interactions, provided that it is able to correctly profit the user feedback. On the other hand, the number of mouse actions may be increased in character-level systems, since the user can move the mouse along one word for correcting it, spending more than one mouse action. In word-based interaction, words are treated as atomic units; therefore, the number of mouse actions required is potentially lower.

\begin{figure*}[!h]
	\centering	
	\begin{subfigure}[t]{0.49\textwidth}
		\centering
		\centering
\footnotesize

\begin{tikzpicture}[
	every axis/.style={ height=4cm,
					    ybar stacked,
 					    major x tick style=transparent,
					    enlarge x limits=true,
					    width=\textwidth,
					    ymin=0,
					    ymax=65,
					    }
	]
\begin{axis}[bar shift=-6pt,
		  ylabel={KSMR [\%]},
		  symbolic x coords={XRCE, UE, UFAL, Europarl, TED}, 		 
	      legend style={
			       	at={(0.2,1.3)},
			       	anchor=north,
			       	legend columns=-1,
	                draw=none},
		  ylabel near ticks,
		  xtick=data,			
		  x tick label style={anchor=east, yshift=-3pt, rotate=45, align=center}, 
			]

	% Word level interactions
	\addplot+[black!60, fill=red!25,
	postaction={pattern=crosshatch dots, pattern color=black!60}
	] coordinates  
	{(XRCE, 48.615636271) % Word level keystrokes: 36259 / 74583 * 100
		(UE, 39.224541781) % Word level keystrokes: 53587 / 136616  * 100
		(UFAL, 50.359018041) % Word level keystrokes: 75675 / 150271 * 100
		(Europarl, 51.379201338) % Word level keystrokes: 201538 / 392256 * 100
		(TED,46.178203888) % Word level keystrokes: 78750 / 170535 * 100
	}; 
	\addplot+[black!60, fill=red!5,
	postaction={pattern=crosshatch dots, pattern color=black!60},
	error bars/.cd, 
	y dir=both,
	y explicit,
	error bar style={xshift=-6pt},
	] coordinates
		{(XRCE, 9.187080166)% +-(0,1.2) % Word level mouse actions: 6852 / 74583 * 100
		(UE, 6.487527083)% +-(0,0.8) % Word level mouse actions: 8863 / 136616  * 100
		(UFAL,7.792588058)% +-(0,0.6) % Word level mouse actions: 11710 / 150271 * 100
		(Europarl, 9.865750938)% +-(0,0.4) % Word level mouse actions: 38699 / 392256 * 100
		(TED, 9.618553376)% +-(0,0.6) % Word level mouse aciions: 16403 / 170535 * 100
	}; 
    \legend{Word-level interaction};
\end{axis}
	\begin{axis}[bar shift=6pt,
				 legend style={at={(0.75,1.3)}, 
				 	anchor=north,
				 	legend columns=-1,
	                draw=none},
          		  symbolic x coords={XRCE, UE, UFAL, Europarl, TED}, 		 
  		         xticklabels={,,,,},	
  		         yticklabels={,,}
				 ]
	\addplot+[black!60, fill=Turquoise!25, 
%	pattern = north west lines,
	 error bars/.cd, y dir=both, y explicit] coordinates
		{(XRCE, 21.75049274) % Character-level keystrokes: 16968 / 74583 * 100
		(UE, 12.418750366) % Character-level keystrokes: 16966 / 136616 * 100
		(UFAL, 13.447039016) % Character-level keystrokes: 20207 / 150271 * 100
		(Europarl,20.626070729) % Character-level keystrokes: 80907 / 392256 * 100
		(TED,18.131761808) % Character-level keystrokes: 30921 / 170535 * 100
			}; 
		
		\addplot+[black!60, fill=Turquoise!5,
			error bars/.cd, 
			y dir=both,
			y explicit,
			error bar style={xshift=6pt}] coordinates 
		{(XRCE, 10.56697907) % +-(0,0.7) % Character-level mouse actions: 8627 / 74583 * 100
			(UE, 7.818996311)% +-(0,0.5) % Character-level mouse actions: 10682 / 136616 * 100
			(UFAL, 8.779471754)% +-(0,0.3) % Character-level mouse actions: 13193 / 150271 * 100
			(Europarl, 11.644946158) % +-(0,0.2)% Character-level mouse actions: 45678 / 392256 * 100
			(TED,11.393555575)% +-(0,0.3) % Character-level mouse actions: 19430 / 170535	 * 100	
		}; 
    \legend{Character-level interaction}
	\end{axis} 

\end{tikzpicture}
		\caption{\label{fig:word_level_vs_char_level_task1_ende} En$\rightarrow$De}
	\end{subfigure}%
	~	
	\begin{subfigure}[t]{0.49\textwidth}
		\centering
		\centering
\footnotesize
\begin{tikzpicture}[
every axis/.style={ height=4cm,
	ybar stacked,
	major x tick style=transparent,
	enlarge x limits=true,
	width=\textwidth,
	ymin=0,
	ymax=60,
}
]
\begin{axis}[bar shift=-6pt,
ylabel={KSMR [\%]},
symbolic x coords={XRCE, UE, UFAL, Europarl, TED}, 		 
legend style={
    at={(0.2,1.3)},
	anchor=north,
	legend columns=-1,
	draw=none},
ylabel near ticks,
xtick=data,			
x tick label style={anchor=east, yshift=-3pt, rotate=45, align=center}, 
]

% Word level interactions
\addplot+[black!60, fill=red!25,
postaction={pattern=crosshatch dots, pattern color=black!60}
%	 error bars/.cd, y dir=both, y explicit
] coordinates  
{(XRCE, 39.579749689) % Word level keystrokes: 26691 / 67436 * 100
	(UE, 23.9881059) % Word level keystrokes:  33479 / 139565  * 100
	(UFAL, 30.09985463) % Word level keystrokes:  47416 / 157529 * 100
	(Europarl,39.867705134) % Word level keystrokes: 181717 / 455800 * 100
	(TED, 36.738771409) % Word level keystrokes: 63622 / 173174 * 100
}; 
\addplot+[black!60, fill=red!5,
postaction={pattern=crosshatch dots, pattern color=black!60}
%	postaction={pattern=dots}, 
%	error bars/.cd, y dir=both, y explicit
] coordinates
{(XRCE,9.259149416 ) % Word level mouse actions: 6244 / 67436 * 100
	(UE,5.458388564) % Word level mouse actions:  7618 / 139565  * 100
	(UFAL, 5.988738581) % Word level mouse actions: 9434 / 157529 * 100
	(Europarl,9.086660816 ) % Word level mouse actions: 41417 / 455800 * 100
	(TED,9.389400256) % Word level mouse aciions: 16260 / 173174 * 100
}; 

\legend{Word-level interaction}
\end{axis}

\begin{axis}[bar shift=6pt,
legend style={at={(0.75,1.3)},
	anchor=north,
	legend columns=-1,
	draw=none},
symbolic x coords={XRCE, UE, UFAL, Europarl, TED}, 		 
xticklabels={,,,,},	
yticklabels={,,}
]

\addplot+[black!60, fill=Turquoise!25, 
] coordinates
{(XRCE, 17.894952251) % Character-level keystrokes: 12742 / 67436 * 100
	(UE, 9.947336367) % Character-level keystrokes: 13883 / 139565 * 100
	(UFAL,12.827479385) % Character-level keystrokes: 20207  / 157529 * 100
	(Europarl, 18.541026766) % Character-level keystrokes: 84510 / 455800 * 100
	(TED,16.619122963) % Character-level keystrokes: 28780 / 173174 * 100
}; 

\addplot+[black!60, fill=Turquoise!5,
pattern color = Turquoise,
%		  pattern = north west lines, 
error bars/.cd, y dir=both, y explicit] coordinates 
{(XRCE,9.040631117) % Character-level mouse actions:  6771 / 67436 * 100
	(UE, 6.390570702) % Character-level mouse actions:  8919 / 139565 * 100
	(UFAL, 7.385941636) % Character-level mouse actions: 11635 / 157529 * 100
	(Europarl, 10.307591049) % Character-level mouse actions: 46982 / 455800 * 100
	(TED, 10.356058069) % Character-level mouse actions: 17934 / 173174		 * 100
}; 
\legend{Character-level interaction}
\end{axis} 

\end{tikzpicture}
		\caption{\label{fig:word_level_vs_char_level_task1_enfr} En$\rightarrow$Fr}
	\end{subfigure}
	\caption{KSMR of INMT systems of all tasks. We compare word-level interaction (dotted) versus character-level interaction. From each bar, the upper (lighter) part represents the mouse action fraction of KSMR, and the lower part accounts for the keystrokes. \label{fig:word_level_vs_char_level_task1}}
\end{figure*}
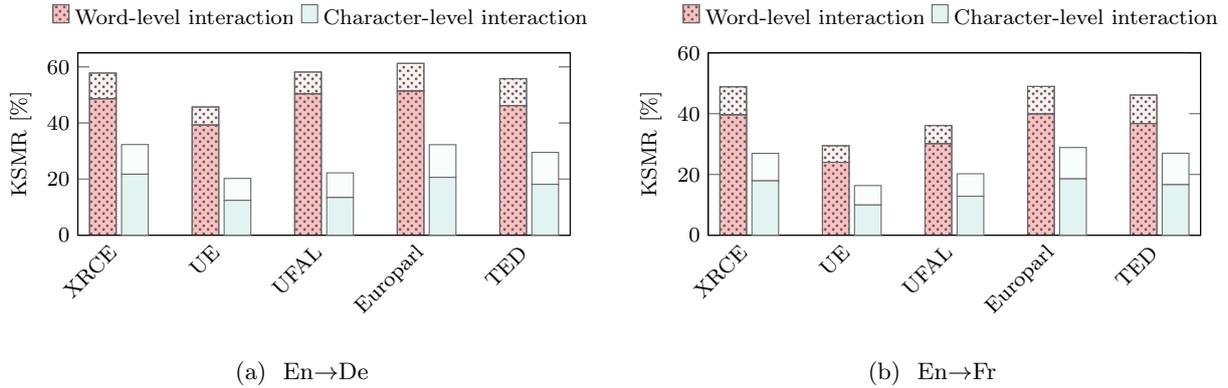

In order to assess the proposed character-based INMT systems, we measured the KSMR required by the same INMT system when performing interactions at either word or character level. Results are shown in \cref{fig:word_level_vs_char_level_task1}. The INMT systems are those from~\cref{table:results-task1-literature}. From KSMR, we differentiated keystrokes and mouse actions.

According to \cref{fig:word_level_vs_char_level_task1}, to perform character level interactions greatly diminished the number of keystrokes required. Reductions were around $50\%$ in the case of French and even larger in the case of German (from a $60\%$ to a $75\%$). This suggests that the system was able to correctly predict even the long and compounded words from German.
As expected, character-based interaction slightly rose the amount of mouse actions required. Nevertheless, the increase of mouse actions was small.

Comparing both levels of interactions, conclusions are indisputable: to perform character-level interaction is more effective than the word-level one, in terms of the human effort required. Moreover, character-level interaction allows the user to have a more precise and natural control of the IMT process.

\subsubsection{Qualitative analysis}

We show an example of a real INMT session, using static and online systems. The NMT system was trained only with in-domain data (scenario \#1) and the sentence belongs to the UFAL (En$\rightarrow$De) task.

%Sentence 916 UFAL EnDe 
\begin{figure}[!h]
	\centering
	\def\arraystretch{1.4}
		\tiny
		\begin{tabular}{ccl}
			\toprule
			\multicolumn{2}{l}{\textbf{Source ($x$)}:}& What is the safe and effective route and duration of antibiotic treatment for children with acute pyelonephritis ? \\
			\multicolumn{2}{l}{\textbf{Target ($\hat{y}$)}:}&  Was ist die sichere und effektive Methode und Dauer einer Antibiotikabehandlung bei Kindern mit akuter Pyelonephritis ?\\
			\hline
			\textbf{IT-0} & \textbf{MT} & Wie sicher und wirksam und wirkdauer für Kinder mit akuter Pyelonephritis ? \\
			\midrule
			\multirow{2}{*}{\textbf{IT-1}} & $\bf{User}$& \textit{\color{darkgreen}{W}}\regularbox{a}ie sicher und wirksam und Wirkdauer für Kinder mit akuter Pyelonephritis ?\\
										   & $\bf{MT}$ & \textit{\color{darkgreen}{Wa}}s ist die sichere und wirksame Art und Dauer der Behandlung von Kindern mit akuter Pyelonephritis ? \\
			\midrule
			\multirow{2}{*}{\textbf{IT-2}} & $\bf{User}$& \textit{\color{darkgreen}{Was ist die sichere und }}\regularbox{e}wirksame Art und Dauer der Behandlung von Lindern mit akuter Pyelonephritis ? \\
										   & $\bf{MT}$ & \textit{\color{darkgreen}{Was ist die sichere und e}}ffektive Art und Dauer der Behandlung von Kindern mit akuter Pyelonephritis ? \\
			\midrule
			\multirow{2}{*}{\textbf{IT-3}} & $\bf{User}$ & \textit{\color{darkgreen}{Was ist die sichere und effektive }}\regularbox{M}art und Dauer der Behandlung von Kindern mit akuter Pyelonephritis ? \\
										   & $\bf{MT}$ & \textit{\color{darkgreen}{Was ist die sichere und effektive M}}enge und Dauer der Antibiotischen Behandlung bei Kindern mit akuter Pyelonephritis ?\\
			\midrule
			\multirow{2}{*}{\textbf{IT-4}} & $\bf{User}$ & \textit{\color{darkgreen}{Was ist die sichere und effektive Me}}\regularbox{t}nge und Dauer der Antibiotischen Behandlung bei Kindern mit akuter Pyelonephritis ?\\
										   & $\bf{MT}$  & \textit{\color{darkgreen}{Was ist die sichere und effektive Met}}hode und Dauer der Antibiotischen Behandlung bei Kindern mit akuter Pyelonephritis ? \\
			\midrule						
			\multirow{2}{*}{\textbf{IT-5}} & $\bf{User}$ & \textit{\color{darkgreen}{Was ist die sichere und effektive Methode und Dauer }}\regularbox{e}der Antibiotischen Behandlung bei Kindern mit akuter Pyelonephritis ? \\
										   & $\bf{MT}$  & \textit{\color{darkgreen}{Was ist die sichere und effektive Methode und Dauer e}}iner Antibiotischen Behandlung bei Kindern mit akuter Pyelonephritis ?\\
			\midrule
			\multirow{2}{*}{\textbf{IT-6}} & $\bf{User}$ & \textit{\color{darkgreen}{Was ist die sichere und effektive Methode und Dauer einer Antibioti}}\regularbox{k}schen Behandlung bei Kindern mit akuter Pyelonephritis ?\\
										   & $\bf{MT}$  &  \textit{\color{darkgreen}{Was ist die sichere und effektive Methode und Dauer einer Antibioti}}kabehandlung bei Kindern mit akuter Pyelonephritis ? \\
			\midrule
			\textbf{END} & $\bf{User}$ &\textit{\color{darkgreen}{ Was ist die sichere und effektive Methode und Dauer einer Antibiotikabehandlung bei Kindern mit akuter Pyelonephritis ?}}\\
			
			\bottomrule
		\end{tabular}

	\caption{\label{fig:offline-INMT} Real INMT session from the UFAL task (scenario \#1). \textbf{IT-} refers to the number of iteration of the process, the \textbf{MT} row refers to the INMT hypothesis in the current iteration and in the \textbf{User} row is shown the feedback introduced by the user: the correct character (boxed). We color in green the prefix that the user has inherently validated while introducing the correction. 12 user actions are required, involving 6 keystrokes and 6 mouse actions (counting final hypothesis acceptation). This represents a KSMR of $10.0$\%. %The regular post-editing of the initial hypothesis requires 10 edit operations (TER=83.3\%).
		}
 \end{figure}

The source sentence is ``\textit{What is the safe and effective route and duration of antibiotic treatment for children with acute pyelonephritis ?}'' and the desired translation is ``\textit{Was ist die sichere und effektive Methode und Dauer einer Antibiotikabehandlung bei Kindern mit akuter Pyelonephritis ?}''.  The static NMT system proposed the translation ``\textit{Wie sicher und wirksam und wirkdauer für Kinder mit akuter Pyelonephritis ?}'', which contains several mistakes. \cref{fig:offline-INMT} shows the corresponding INMT session for this example. In this case, 6 iterations were required, in order to match the desired translation. 
 
It is interesting to deepen on how the system reacted to the feedback provided. In the first iteration, the user introduced a correction in the interrogative word ``\textit{Wie}'' (``How''). The system not only correctly predicted the new interrogative word, ``\textit{Was}'' (``What''), but also changed the hypothesis, matching the correct hypothesis. It is also remarkable that the system naturally handled compound words: in the sixth iteration, the system transformed the words ``\textit{Antibiotischen Behandlung}'' into the compound ``\textit{Antibiotikabehandlung}'' from the user keystroke \regularbox{k}.

\cref{fig:online-INMT} shows the same INMT session, but for an adaptive NMT system. Previously to this sample, the system was already adapted with 915 sentences. The initial hypothesis was better than one proposed by the static system. E.g: the initial interrogative clause was correctly translated. Moreover, it reacted better to the user feedback: in the second iteration, the system correctly predicted the word ``\textit{Methode}'' with a single keystroke.

\begin{figure}[!h]
	\centering
	\def\arraystretch{1.4}
	\tiny
	\begin{tabular}{ccl}
		\toprule
		\multicolumn{2}{l}{\textbf{Source ($x$)}:}& What is the safe and effective route and duration of antibiotic treatment for children with acute Pyelonephritis ? \\
		\multicolumn{2}{l}{\textbf{Target ($\hat{y}$)}:}&  Was ist die sichere und effektive Methode und Dauer einer Antibiotikabehandlung bei Kindern mit akuter Pyelonephritis ?\\
		\hline
		\textbf{IT-0} & \textbf{MT} & Was ist die sichere und wirksame Art und Dauer der Antibiotischen Behandlung bei Kindern mit akuter Pyelonephritis ?\\
		\midrule
		\multirow{2}{*}{\textbf{IT-1}} & $\bf{User}$& \textit{\color{darkgreen}{Was ist die sichere und }}\regularbox{e}wirksame Art und Dauer der Antibiotischen Behandlung bei Kindern mit akuter pyelonephritis ?\\
								    	& $\bf{MT}$ & \textit{\color{darkgreen}{Was ist die sichere und e}}ffektive Art und Dauer der Antibiotischen Behandlung bei Kindern mit akuter Pyelonephritis ?\\
		\midrule
		\multirow{2}{*}{\textbf{IT-2}} & $\bf{User}$& \textit{\color{darkgreen}{Was ist die sichere und effektive }}\regularbox{M}Art und Dauer der Antibiotischen Behandlung bei Kindern mit akuter pyelonephritis ?\\
										& $\bf{MT}$ & \textit{\color{darkgreen}{Was ist die sichere und effektive M}}ethode und Dauer der Antibiotischen Behandlung bei Kindern mit akuter Pyelonephritis ?\\
		\midrule
		\multirow{2}{*}{\textbf{IT-3}} & $\bf{User}$ & \textit{\color{darkgreen}{Was ist die sichere und effektive Methode und Dauer }}\regularbox{e}der Antibiotischen Behandlung bei Kindern mit akuter pyelonephritis ?\\
								   		& $\bf{MT}$  & \textit{\color{darkgreen}{Was ist die sichere und effektive Methode und Dauer e}}iner Antibiotischen Behandlung bei Kindern mit akuter Pyelonephritis ?\\
		\midrule
		\multirow{2}{*}{\textbf{IT-4}} & $\bf{User}$ & \textit{\color{darkgreen}{Was ist die sichere und effektive Methode und Dauer einer antibioti}}\regularbox{k}schen Behandlung bei Kindern mit akuter Pyelonephritis ? \\
										& $\bf{MT}$  & \textit{\color{darkgreen}{Was ist die sichere und effektive Methode und Dauer einer Antibiotik}}abehandlung bei Kindern mit akuter Pyelonephritis ? \\
		\midrule
		\textbf{END} & $\bf{User}$ &\textit{\color{darkgreen}{Was ist die sichere und effektive Methode und Dauer einer Antibiotikabehandlung bei Kindern mit akuter Pyelonephritis ?}}\\
		
		\bottomrule
	\end{tabular}

\caption{\label{fig:online-INMT} Same IMT session and notation as \cref{fig:offline-INMT} but incorporating OL. Only 4 keystrokes and are 5 mouse actions are now required (KSMR=$7.6\%$).
}
\end{figure}

\section{Conclusions and future work}
\label{sec:Conclusions}

In this work, we studied the application of online learning techniques to NMT. For building our NMT systems, we introduced the cLSTM unit, an extension of the successful cGRU to the LSTM architecture. It consists in the cascaded application of two LSTM blocks in the decoder, with an attention model in between. We presented preliminary results that show that this modification of the LSTM cell offers better performance than standard LSTM units. However, to perform an exhaustive comparison between them is out of the scope of this work and we leave it to future investigations. 

We tested our proposals in two computer assisted translation frameworks: post-editing and interactive machine translation. We introduced a novel and effective way for interacting at a character-level with a word-level NMT system. We assessed our strategy and found that it approximately halved the KSMR required during the IMT process.

We studied the application of these systems in three different scenarios. All of them referred to plausible situations in the translation industry and relate to the amount of training data available: we may have enough in-domain data to properly train a NMT system, to also have an out-of-domain corpus to provide additional knowledge to the NMT or we can suffer from lack of in-domain data.  

We conducted a wide experimentation, relating two language pairs in five different domains, for each one of the proposed scenarios. The results were conclusive: online learning techniques were able to bring significant improvements to static systems in almost every case. Adaptive NMT systems produced better translation hypotheses and reduced the human effort required for correcting their outputs. The magnitude of such enhancements were task-dependent, according to the properties of the text to translate. We also faced a discouraging case in which, due to the features of the data, OL was ineffective. We also should investigate solutions to these situations.

The application of online learning to INMT systems reduced even more the human effort required for correcting translation hypotheses. Moreover, the computational overhead of OL was small, making suitable the use of adaptive NMT systems in an interactive scenario. We also compared our system with the state-of-the-art in IMT, based on PB-SMT models. Neural systems beat PB-SMT by a large margin in terms of the human effort required.

As future work, we aim to boost the effectiveness of online learning for NMT. A main issue suffered by SGD is that the objective function (target sentence likelihood) is not necessarily correlated with the assessment criteria (TER, BLEU, KSMR). Therefore, to optimize the likelihood may be suboptimal for the aforementioned metrics. Reinforcement learning has been used for directly optimizing BLEU, as a complementary method to the traditional maximum likelihood training \citep{Wu16} or during decoding \citep{Gu17}. The application of reinforcement learning for adaptive NMT \citep{Kreutzer17} is also a promising research direction.

Tightly related to online learning and IMT is the active learning field \citep{Olsson09}. Under this paradigm, the system has available a large pool of unlabeled instances and interactively queries to an oracle (e.g. a human agent) to label some of them. It is especially adequate for situations in which the manual labeling is expensive, as machine translation. The active learning framework has brought interesting gains when combined with INMT \citep{Lam18,Peris18c}. To further investigate towards this direction seems also encouraging.

Finally, it is mandatory to test our interactive, adaptive systems with real users. We have this point at the top of our agenda and we hope to perform a human evaluation in a near future.

\section*{Acknowledgements}

The authors wish to thank the anonymous reviewers for their valuable criticisms and suggestions. The research leading to these results has received funding from
the Generalitat Valenciana under grant PROMETEOII/2014/030 and from TIN2015-70924-C2-1-R.

\section*{References}
\bibliographystyle{elsarticle-harv}
\bibliography{0_main}

\end{document}